\documentclass{article}

\usepackage{PRIMEarxiv}

\usepackage[utf8]{inputenc} % allow utf-8 input
\usepackage[T1]{fontenc}    % use 8-bit T1 fonts
\usepackage{hyperref}       % hyperlinks
\usepackage{url}            % simple URL typesetting
\usepackage{booktabs}       % professional-quality tables
\usepackage{amsfonts}       % blackboard math symbols
\usepackage{nicefrac}       % compact symbols for 1/2, etc.
\usepackage{microtype}      % microtypography
\usepackage{lipsum}
\usepackage{fancyhdr}       % header
\usepackage{graphicx}       % graphics
\graphicspath{{media/}}     % organize your images and other figures under media/ folder

\usepackage{subfig}
\usepackage[numbers,sort&compress]{natbib}
\usepackage{amsmath,amssymb,amsthm}
\usepackage{mathtools}
\usepackage[ruled]{algorithm2e}

\title{Blindly Deconvolving\\Super-noisy Blurry Image Sequences
}

\author{
  Leonid Kostrykin \\
  Biomedical Computer Vision Group, \\
  BioQuant, IPMB, Heidelberg University \\
  \texttt{leonid.kostrykin@bioquant.uni-heidelberg.de} \\
   \And
  Stefan Harmeling \\
  Department of Computer Science \\
  Technical University Dortmund \\
  \texttt{stefan.harmeling@tu-dortmund.de} \\
}

\makeatletter
\g@addto@macro\bfseries{\boldmath}
\makeatother

\newcommand{\inline}[1]{\raisebox{0pt}[0pt][0pt]{#1}}

\newcommand{\eqrefrel}[2]{\stackrel{\rlap{\scriptsize\eqref{#1}}}{#2}}

\newcommand{\cnvmat}[1]{#1}

\newcommand{\shape}[1]{N_{#1}}

\newcommand{\conj}[1]{\overline{#1}}

\usepackage[table]{xcolor}

\SetCommentSty{}
\SetAlgoInsideSkip{smallskip}

\numberwithin{equation}{section}

\newcommand{\T}{\mathsf{T}}
\newcommand{\Hermitian}{\mathsf{H}}

\newcommand{\differential}{\mathrm{d}}
\newcommand{\trace}{\operatorname{tr}}
\newcommand{\rank}{\operatorname{rk}}

\newcommand{\Diag}{\operatorname{Diag}}
\newcommand{\SPAN}{\operatorname{span}}
\newcommand{\range}{\operatorname{range}}

\DeclareMathOperator{\Cov}{Cov}

\AtBeginDocument{%
}

\begin{document}
\maketitle

\begin{abstract}
Image blur and image noise are imaging artifacts intrinsically arising in image acquisition. In this paper, we consider \emph{multi-frame blind deconvolution} (MFBD), where image blur is described by the convolution of an unobservable, undeteriorated image and an unknown filter, and the objective is to recover the undeteriorated image from a sequence of its blurry and noisy observations. We present two new methods for MFBD, which, in contrast to previous work, do not require the estimation of the unknown filters.

The first method is based on likelihood maximization and requires careful initialization to cope with the non-convexity of the loss function. The second method circumvents this requirement and exploits that the solution of likelihood maximization emerges as an \emph{eigenvector} of a specifically constructed matrix, if the \emph{signal subspace} spanned by the observations has a sufficiently large dimension.

We describe a pre-processing step, which increases the dimension of the signal subspace by artificially generating additional observations. We also propose an extension of the eigenvector method, which copes with insufficient dimensions of the signal subspace by estimating a \emph{footprint} of the unknown filters (that is a vector of the size of the filters, only one is required for the whole image sequence).

We have applied the eigenvector method to synthetically generated image sequences and performed a quantitative comparison with a previous method, obtaining strongly improved results.
\end{abstract}

% keywords can be removed

\section{Introduction}
\label{chap:introduction}

\begin{figure}[h!]
	\centering
	\subfloat[Undeteriorated observation]{
		\includegraphics[height=40mm]{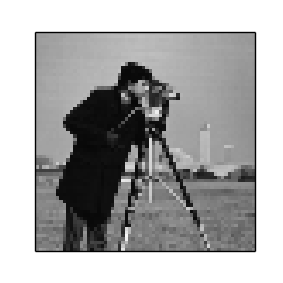}
    	\label{fig:intro:cameraman}}%
	\subfloat[Blurry observation]{
		\includegraphics[height=40mm]{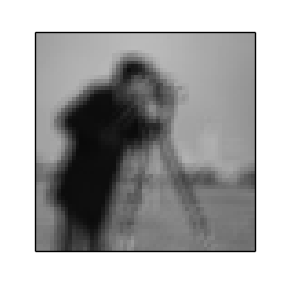}
    	\label{fig:intro:y:noise_free}}%
	\subfloat[Blurry and noisy observation]{
		\includegraphics[height=40mm]{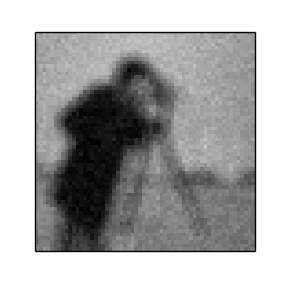}
    	\label{fig:intro:y:noisy}}
	\caption{Common imaging artifacts (synthetically generated).}
	\label{fig:intro}
\end{figure}

Image acquisition is an indispensable step in many technical applications, including digital photography (e.g., \cite{brown2007automatic}), visual inspection of industrial devices and other structures (e.g., \cite{kostrykin2021,gui2020automated,yang2019deep}), surveillance (e.g., \cite{ramaswamy2018frame,cheng2009accurate,senarathne2011faster}), and biomedical image analysis (e.g, \cite{kostrykin2022,horl2019bigstitcher,stringer2020cellpose}). Despite of a broad range of applications, the acquisition of images is an error-prone task: Challenging imaging conditions like imperfections of the optical systems, camera shake, low-light conditions, or high relative velocities are common causes of imaging artifacts which are perceived as blurriness. In addition, the acquired or \emph{observed} image data is often deteriorated by image noise (see \autoref{fig:intro}). In some application areas such as digital photography, noise can be suppressed by longer exposures, which, however, comes at the cost of increased blurriness. On the opposite, shorter exposures tend to yield observations which are less blurry, but noisier. So, at the end of the day, image restoration techniques are required.

This paper is widely based on the unpublished master's project of \citet{kostrykin2016_thesis}. Throughout the paper, we assume that the image blur is invariant w.r.t.\ the location within an image. Using this assumption, it is convenient to describe the blur of an observed image by the \emph{convolution} $x \ast a$ of an unobservable and undeteriorated \emph{ground truth} $x$ and some \emph{filter} $a$, where the latter characterizes the blur. Our work can be extended to tackle spatially variant blur using \cite{hirsch2010}.

The recovery of the undeteriorated image $x$ from its blurry \emph{observation} $y$ is called \emph{deconvolution}, where it is $y = x \ast a$, in the noise-free case. In practical applications, not only $x$, but also the filter $a$ is unknown, which is referred to as \emph{blind} deconvolution. The blind deconvolution problem is ill-posed: For instance, $y = x$ with $a$ being the identity element of convolution solves the linear system $y = a \ast x$ for any observation $y$ (but this is not a meaningful solution).

The blind deconvolution problem becomes tractable, either by making prior assumptions regarding $a$ and $x$, or by taking more than just one observation of the same $x$ into account. The recovery of $x$ from a sequence of $n$ observations $y_1, \dots, y_n$ is called \emph{multi-frame} blind deconvolution (MFBD). We will write $a_i$ to denote the filter, which characterizes the $i$-th observation $y_i$ of $x$. Note that displacements of the observed object w.r.t.\ the imaging system within the image sequence are tolerable since translations can be expressed by convolution. In this paper, we consider the case that the observations are very noisy, but also $n$ is very large. Such a setting is common in, for example, astronomical imaging.

\subsection{Notational conventions}
\label{chap:notation}

We use the following notation throughout this paper.

\textbf{Functions.} For notation of function values, we write square brackets to indicate that the domain of a function is discrete, and round brackets to indicate that it is continuous. We also implicitly consider functions with discrete domain and finite support as column vectors (tuples). Consequently, we also use square brackets to represent components of vectors.

\textbf{Matrices and vectors.} For explicit notation of vectors and matrices, we use square brackets like $\left[ y_1\, \dots\, y_n \right]$ to write a matrix consisting of the columns $y_1, \dots, y_n$, and we use round brackets like $\left(\alpha_1, \dots, \alpha_n\right)$ to write a tuple consisting of the components $\alpha_1, \dots, \alpha_n$ (i.e.\ a column vector). Superscript $\T$ stands for transposition and $\left\|\cdot\right\|$ implies the $\ell_2$ norm.

\textbf{Probabilities.} We write $p\left(y \middle| x, a\right)$ to denote the conditional probability of $y$, given $x$ and $a$. 

\subsection{Blur through discrete convolution}
\label{chap:background}

Although the blurring in image acquisition takes place before the continuous image signal is discretized by the digital image sensor, and is thus subject to \emph{continuous} mechanics, it nevertheless can be modeled through \emph{discrete} convolution (for justification, see Section~B3.2 in \cite{bremaud2002}). Discrete convolution is a commutative operation, that takes two images represented by functions $g, h$ and yields a new one. One of the two input images is called the \emph{filter}. However, due the commutativity, the naming is context-dependent. Formally, we rely on the definition from \cite{mcclellan2003} for discrete convolution,
\begin{gather}
	\label{eq:convolution1d}
	\left(g \ast h\right)\left[t\right] =
		\sum_{k=-\infty}^\infty g\left[t-k\right] \cdot h\left[k\right]
	\text,
\end{gather}
with two functions $g, h\colon \mathbb Z \to \mathbb R$. To simplify notation, we will only describe the one-dimensional case, whenever the two-dimensional case behaves analogously. Otherwise, the differences will be pointed out.

Given that the functions $g,h$ represent digital images, it is plausible to assume that they have finite support. This motivates considering $g$ and $h$ as vectors of dimensions $\shape{g}$ and $\shape{h}$, respectively. For algebraic considerations, we will implicitly consider $g,h$ as vectors, obtained by column-wise concatenation of the two-dimensional images which they represent. For an undeteriorated image $x$ and a filter $a$, we see from Eq.~\eqref{eq:convolution1d} that the convolution $x \ast a$ is linear in both, $x$ and $a$. Thus, convolution can be written as the \emph{matrix-vector-product} $a \ast x = \cnvmat Ax$, but also $a \ast x = \cnvmat Xa$, where the matrices $\cnvmat A$ and $\cnvmat X$ are induced by the vectors $a$ and $x$, respectively (this is described in \autoref{chap:theory}).

Throughout this paper, we assume that all filters $a_i \in \mathbb R^{\shape a}$ are of equal size $\shape a$, and particularly smaller than the image $x \in \mathbb R^{\shape x}$ in every dimension, which we write as $\shape a < \shape x$. We say, that the filter $a$ is a \emph{point spread function} (PSF), if it neither has negative elements, nor its application to an image changes the image's brightness, i.e.\ $\sum_k a\left[k\right] = 1$.

\subsection{Previous approaches}
\label{chap:related_work}

Early work on deconvolution of noisy images included \cite{richardson1972bayesian,lucy1974iterative}. \citet{richardson1972bayesian} derived multiplicative updates for the case of non-blind deconvolution. It was assumed that the object and the image are probability distributions on the pixels. Bayes’ formula together with the definition of conditional probability leads to the update formula. Curiously, the conditional probability of an image pixel given an object pixel is the PSF. A similar derivation was performed by \citet{lucy1974iterative}, who also showed how this can be seen as an approximation of likelihood maximization. We refer the reader to \cite{tong1998multichannel} for a comprehensive overview of the early work.

A prominent approach specifically for \emph{multi-frame} blind deconvolution (MFBD) was proposed by \citet{harikumar1999}. The authors used the \emph{likelihood maximization} approach $\hat a = \arg\max_a \max_x p\left(y \middle| x, a\right)$, where $y = \left[y_1\, \dots\, y_n\right]$ and $a = \left[a_1\, \dots\, a_n\right]$, and derived the estimate $\hat{a} = \beta \cdot \arg\min_a a^\T R\, a$ of the filters, where $\beta = \left\| \hat a \right\|$ is a positive factor and the matrix $R$ can be constructed from the true and unknown filters $a$. They constrained that the norm of $\arg\min_a a^\T R\, a$ should be $1$ to avoid the trivial solution $\hat a = 0$ and showed that $\pm \hat a / \beta$ are the eigenvectors of $R$, which correspond to its smallest eigenvalue. To determine these eigenvectors, the authors used an approximation of the matrix $R$ which was refined iteratively. It is easy to recover the norm of each filter $\hat{a}_i$ using the assumption that it is a PSF. However, each iteration for the refinement of the estimated matrix $R$ requires solving a least squares problem that involves the whole sequence of observations, rendering the method infeasible for large $n$.

\citet{sroubek2012} observed that the method of \citet{harikumar1999} fails in the presence of noise and addressed this issue via regularization. They employed a prior for $x$ which favors a sparse gradient. For the regularization of the filters $a$, they derived another matrix $R_\Delta$ so that $\hat{a} \approx \arg\min_a a^\T R_\Delta a$; but in contrast to the work of \cite{harikumar1999}, their $R_\Delta$ does not depend on the unknown filters $a$ and is robust to noise. They minimized the resulting objective function w.r.t.\ $x$ and $a$ alternatingly. However, the $n\,\shape a \times n\,\shape a$ matrix $R_\Delta$ is dense, which makes the method impractical for large $n$.

To cope with large image sequences, \citet{harmeling2009} proposed an online algorithm for MFBD, which considers only a single observation per iteration. The authors used the loss function \inline{$\sum_i^n \left\|y - x \ast a_i\right\|^2$}, whose minimization is equivalent to the maximization of the likelihood $p\left(y \middle| x, a\right)$ under mild conditions, and derived multiplicative updates for $x$ and $a$.

The abovementioned methods have in common that the filters $a$ are determined alongside, although only the undeteriorated image $x$ is of interest. This means that the parameter space is larger than required (a filter needs to be estimated for each image of the sequence, which is a potentially very large number), causing additional computational cost. To the best of our knowledge, this concerns all previously developed methods for MFBD.

\subsection{Contributions}
\label{chap:contributions}

In this paper, we propose two methods which eliminate the need for estimating the filters $a$ in order to determine the undeteriorated image $x$. This not only has the advantage that fewer variables must be computed, but also that, in the special cases described below, the undeteriorated image $x$ can be determined without alternating optimization schemes.

In our work, the estimate $\hat x_\ast$ of the undeteriorated image $x$ appears as an eigenvector of a specific matrix. We show, that this matrix is fully determined solely by the observations, and its eigenvector $\hat x_\ast$ maximizes the likelihood of the observations, when two specific conditions are met:

\begin{enumerate}
	\item The first condition is that the observations are noise-free -- however, we argue that this condition is also attained asymptotically in the presence of noise, if a sufficiently large number $n$ of observations is taken into account. This is achieved by a subspace technique \citep{moulines1995}, that makes the proposed methods cope with any noise level as long as $n$ is sufficiently large.

	\item The second condition concerns the filters which characterize the blur of the observed images and requires that they span a space of a sufficiently large dimension. We describe a pre-processing step which artificially generates additional observations, making this condition more likely to be fulfilled. Still, this is insufficient in specific cases, and the proposed method falls back to an alternating optimization scheme then.
\end{enumerate}

In \autoref{chap:theory}, we briefly describe the theoretical foundations of our work. In \autoref{chap:approach}, we derive a method based on likelihood maximization which directly determines the undeteriorated image $x$ without estimating the filters $a$. The loss function of this method is non-convex and direct minimization requires reliable initialization. In \autoref{chap:method}, we describe the second method, which exploits that the same solution emerges as an eigenvector of a specific matrix. %The method is concluded by generalization to the case that the second condition does not hold.
The method is applied to synthetically generated image sequences, and the obtained results are compared to those achieved using a previous method \cite{harmeling2009}. Finally, we discuss its advantages and limitations in \autoref{chap:discussion}.

\section{Foundations}
\label{chap:theory}

\subsection{Valid convolution and associativity}
\label{chap:boundary_treatment}

The computation of $x \ast a$ can be done efficiently using the \emph{discrete Fourier transform} (DFT) of $x$ and $a$. Since the computation of the DFT of $x$ (or $a$) is a linear operation, we can write it as the matrix-vector-product $Fx$, where $F$ is the DFT matrix (see, e.g., \cite{kamisetty2001}). We denote the \emph{Hermitian conjugate} of $F$ as \inline{$F^\Hermitian = \conj{F}{\,}^\T$}, where \inline{$\conj F$} for is the complex conjugate of $F$. The matrix $F$ is unitary, i.e.\ $F^\Hermitian = F^{-1}$, thus $F^\Hermitian$ expresses the inverse DFT of the vector to its right.

If $x$ represents an image section from a larger, $\shape x$-periodical image, then the discrete version of the widely known \emph{convolution theorem} (see, e.g., Theorem~B3.2 in \cite{bremaud2002}) states that, the expression
\begin{gather}
	a \ast x = F^\Hermitian \left( F x \odot F I_a a \right)
	\label{eq:convolution:circ}
\end{gather}
is a period of the periodical, convolved image. The matrix $I_a$ pads the vector to its right with zeros and $\odot$ stands for element-wise multiplication. The computation of $a \ast x$ through Eq.~\eqref{eq:convolution:circ}, which is called \emph{circular convolution}, causes artifacts at the boundaries when used for non-periodic signals.

To cope with that, we only keep the \emph{valid} section of $a \ast x$ computed by Eq.~\eqref{eq:convolution:circ}, which is the section where the periodicity artifacts do not occur. This is referred to as \emph{valid convolution} and we write
\begin{align}
    a \ast_\text{valid} x =
	    I_y^\T \left(a \ast x\right)
	\eqrefrel{eq:convolution:circ}{=}
    	I_y^\T F^\Hermitian \left( F x \odot F I_a a \right)
    \text,
    \label{eq:convolution:valid}
\end{align}
where the matrix $I_y^\T$ crops $a \ast x$ so that its size equals $\shape y = \shape x - \shape a + 1$. Using Eq.~\eqref{eq:convolution:valid}, we can write the matrix-vector-products $Ax$ and $Xa$ using the matrices $A$ and $X$ defined as functions of $a$ and $x$,
\begin{align}
    \cnvmat A\!\left(a\right) &= I_y^\T F^\Hermitian \Diag \left( F I_a a \right) F, \\
    \cnvmat X\!\left(x\right) &= I_y^\T F^\Hermitian \Diag \left( F x \right) F I_a,
\end{align}
respectively. This directly leads to the curious rule of associativity
\begin{gather}
	b \ast_\text{valid} \left(a \ast_\text{valid} x\right)
	=	\left(b \ast_\text{full} a\right) \ast_\text{valid} x
	\text,
    \label{eq:convolution:associativity}
\end{gather}
where $b \in \mathbb R^{\shape b}$ is another filter and
\begin{gather}
	b \ast_\text{full} a =
    	F^\Hermitian \left( F I_b b \odot F I_a a \right)
    \label{eq:convolution:full}
\end{gather}
denotes \emph{full convolution}, a different way of avoiding periodicity artifacts. The matrices $I_b$ and $I_a$ zero-pad $b$ and $a$ to the size of the result, that is $\shape b + \shape a - 1$. Note that valid convolution yields an image smaller than $x$, whereas full convolution yields an image larger than $x$.

\subsection{Multi-frame forward model}
\label{chap:model}

\begin{figure}[htb]
  \begin{center}
  \includegraphics[scale=0.6]{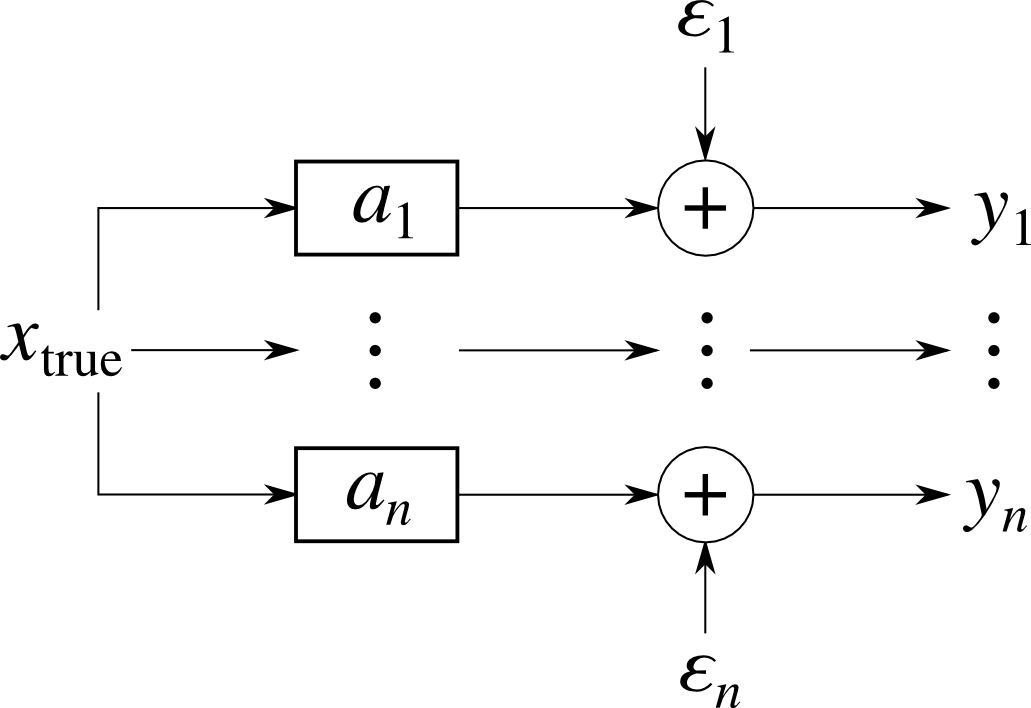}
  \caption{The multi-frame forward model by \citet{harikumar1999}.}
  \label{fig:model}
  \end{center}
\end{figure}

Below, we write $x_\text{true}$ to explicitly denote the unknown and undeteriorated ground truth image (for simplicity, this was denoted by a simple $x$ in \autoref{chap:introduction}). Following \cite{harikumar1999}, we assume that an observation $y_i$ of $x_\text{true}$ is the additive superposition of two unobservable quantities. These are the noise-free, blurry image $\tilde y_i = a_i \ast_\text{valid} x_\text{true}$ and the noise $\varepsilon_i$:
\begin{align}
    \label{eq:model}
    y_i = a_i \ast_\text{valid} x_\text{true} + \varepsilon_i =
    \cnvmat X\!\left(x_\text{true}\right) a_i + \varepsilon_i
\end{align}
\autoref{fig:model} illustrates this modeling. We further assume that $\varepsilon_i$ is \emph{additive white Gaussian noise} (AWGN).

The $\shape y \times \shape a$ matrix $X\!\left(x\right)$ in Eq.~\eqref{eq:model} is structured like
\begin{gather}
	\cnvmat X\!\left(x\right) =
		\left[\begin{array}{cccc}
	    	x_{\shape a} & x_{\shape a - 1} & \cdots & x_1 \\
	    	\vdots  & \vdots      &        & \vdots \\
	    	x_{\shape x} & x_{\shape x - 1} & \cdots & x_{\shape x - \shape a + 1}
	    \end{array}\right]
	\text,
    \label{fig:convolution:valid:Xa}
\end{gather}
that is, the range (column space) of $X$ is spanned by all $\shape y$-sized sections of $x$, which are $\shape a$ in count. This establishes the following intuitive view of Eq.~\eqref{eq:model}: Any noise-free observation of $x$ is a linear combination of all $\shape y$-sized sections of $x$, and $a_i$ corresponds to the weights of the combination (i.e.\ how much each section contributes to the observation).

The dimension $\shape a$ of the filters $a_i$ is a parameter of the model. It controls the number of adjacent pixels, any filter can put into relation at most. Thus, the bigger we choose $\shape a$, the higher the more blur the model is capable to explain.

\subsection{Identifying the signal subspace}
\label{chap:signal_subspace}

Repetitive observation of $x_\text{true}$ yields a sequence of $\shape y$-dimensional vectors according to Eq.~\eqref{eq:model}. In the noise-free case, there are at most $\shape a$ degrees of freedom. Since $y_i$ depends linearly on $a_i$, the vectors $\tilde y_1, \dots, \tilde y_n$ span an $m$-dimensional subspace of $\mathbb R^{\shape y}$, where $m \leq \shape a$ and $\shape a$ may be much smaller than $\shape y$. \citet{moulines1995} called this the \emph{signal subspace}. There are at least three reasons, why we should expect $m < \shape a$:
\begin{enumerate}
	\item The model parameter $\shape a$ might be overestimated.
	\item Natural PSFs aren't rectangular.
	\item Even if variance is encountered in all pixels of the filters, $m < \shape a$ will still hold for filters which are sampled from a \emph{PSF subspace}, i.e.\ a subspace of $\mathbb R^{\shape a}$.
\end{enumerate}
Note that the last reason holds particularly if, but not only if, $n < \shape a$ (i.e.\ the number of observations is too small).

Note that $m$ cannot be larger than the dimension of the PSF subspace. Moreover, as we see from
\begin{gather}
	\rank \left[\tilde y_1\, \dots\, \tilde y_n\right]
	\;\eqrefrel{eq:model}{=}\;
		\rank \left[\cnvmat Xa_1\, \dots\, \cnvmat Xa_n\right]
	\text,
\end{gather}
the dimensions of the signal subspace and the PSF subspace are equal if $\rank \cnvmat X = \shape a$. This property means that none of the $\shape y$-sized sections of $x_\text{true}$ are linearly dependent. \citet{harikumar1999} called those images, which this property holds for, \emph{persistently exciting}. We assume that $x_\text{true}$ is persistently exciting for the rest of this paper.

\subsubsection{Noise-free case}
\label{chap:signal_subspace:noise_free}

\citet{moulines1995} proposed identification of the signal subspace by a process similar to performing PCA (see, e.g., \cite{murphy2012}) without mean subtraction on the noise-free observations $\tilde Y = \left[\tilde y_1\, \dots\, \tilde y_n\right]$. The \emph{eigenvalue decomposition} (EVD)
\begin{gather}
    \frac{1}{n} \tilde Y \tilde Y^\T = U \Lambda U^\T
    \label{eq:noisefree_observations_cov_evd}
\end{gather}
of the empirical covariance matrix $\frac{1}{n} \tilde Y \tilde Y^\T$ of the noise-free observations $\tilde Y$ with non-negative eigenvalues $\Lambda = \operatorname{Diag} \lambda$ induces the matrix $U$. If we put the eigenvalues into descending order $\lambda_i \geq \lambda_{i+1}$, then the first columns of $U$ correspond to the directions with the greatest variance. The signal subspace is then spanned by the $m$ first columns of $U$, and $m$ equals the number of non-zero eigenvalues.

\subsubsection{Noisy case}
\label{chap:signal_subspace:noisy}

If the number $n$ of observations is sufficiently large, then the signal subspace is also identifiable in the presence of noise. To understand this, we will look at how the additive noise vectors $E = \left[\varepsilon_1\, \dots\, \varepsilon_n\right]$ influence the covariance matrix of the noisy observations. Writing $Y = \left[y_1\, \dots\, y_n\right]$, the covariance matrix resolves to
\begin{gather}
	\frac{1}{n} YY^\T =
		\frac{1}{n} \tilde Y \tilde Y^\T +
		\frac{1}{n} \tilde Y E^\T +
		\frac{1}{n} E \tilde Y^\T +
		\frac{1}{n} EE^\T
	\text.
	\label{eq:noisy_observations_cov}
\end{gather}
For $n \to \infty$, the terms $\frac{1}{n} \tilde Y E^\T$ and $\frac{1}{n} E \tilde Y^\T$ both tend to $0$, because \inline{$\tilde Y$} and $E$ are uncorrelated. Furthermore, since we know that the noise \inline{$E \sim \mathcal N\left(0; I \sigma^2\right)$} is white by assumption in \autoref{chap:model}, we conclude that \inline{$\frac{1}{n} EE^\T \xrightarrow{n\to\infty} I \sigma^2$}. Then, plugging the decomposition from Eq.~\eqref{eq:noisefree_observations_cov_evd} into Eq.~\eqref{eq:noisy_observations_cov} yields
\begin{gather}
	\lim_{n\to\infty}
	\frac{1}{n} YY^\T = U \left(\Lambda + \sigma^2 I\right) U^\T
	\text.
	\label{eq:noisy_observations_evd}
\end{gather}
The matrix $\Lambda + \sigma^2 I$ is diagonal. Thus, Eq.~\eqref{eq:noisy_observations_evd} equals the EVD of the covariance matrix of the noisy observations for $n \to \infty$. Notably, the eigenvectors of the EVD are the same as in the noise-free case: This means that for a sufficiently large number of observations, the $m$ first eigenvectors of the covariance matrix $\frac{1}{n} YY^\T$ span the same subspace, as the unobservable noise-free observations $\tilde y_1, \dots, \tilde y_n$.

In practice, we have to rely on a finite number of observations. For any fixed $n$, the matrix \inline{$\frac{1}{n} EE^\T$} deviates the more from diagonal shape, the higher the noise level is. As a consequence, the EVD of \inline{$\frac{1}{n} YY^\T$} also deviates from Eq.~\eqref{eq:noisy_observations_evd}, and we say that the obtained eigenvectors are \emph{misaligned} (w.r.t.\ the ideal eigenvectors of the noise-free observations).

\begin{figure}
    \centering%
	\subfloat[Eigenvalues by noise level]{
		\includegraphics[width=0.38\textwidth]{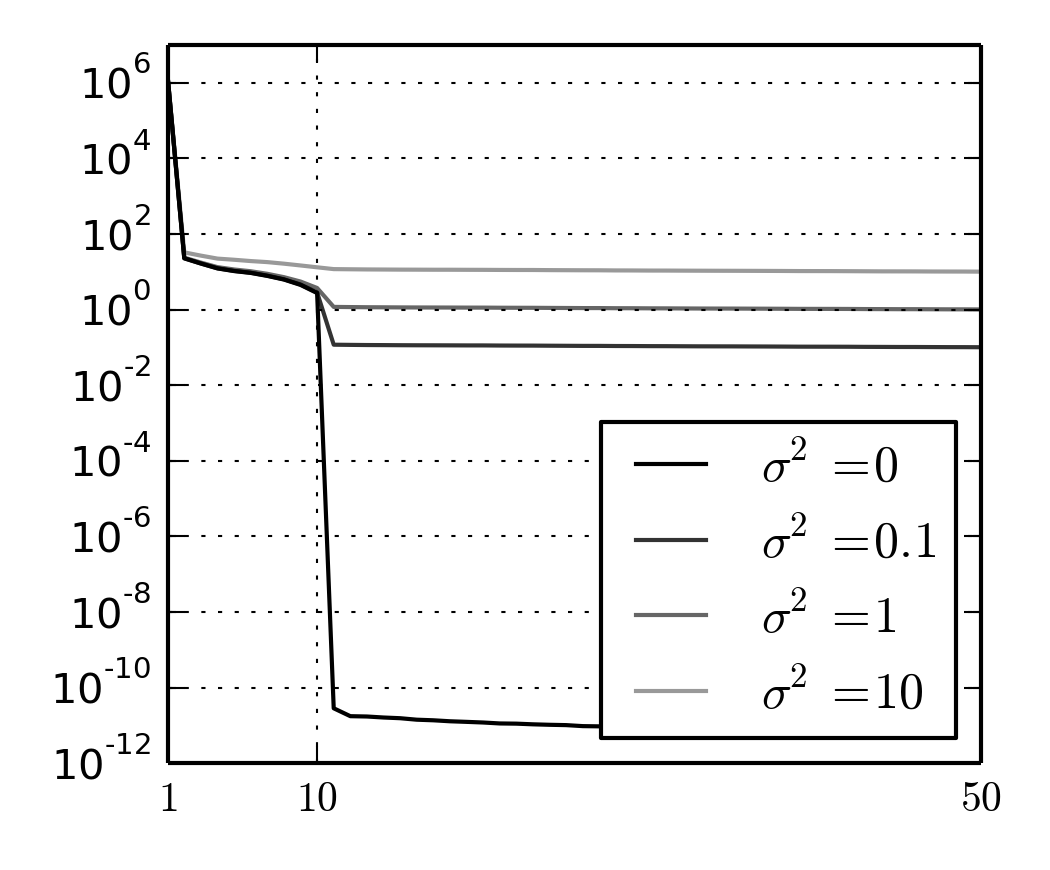}
    	\label{fig:signal_subspace:eigvalues}}
    \quad%
	\subfloat[Misalignment of the eigenvectors (in degrees)]{
		\includegraphics[width=0.38\textwidth]{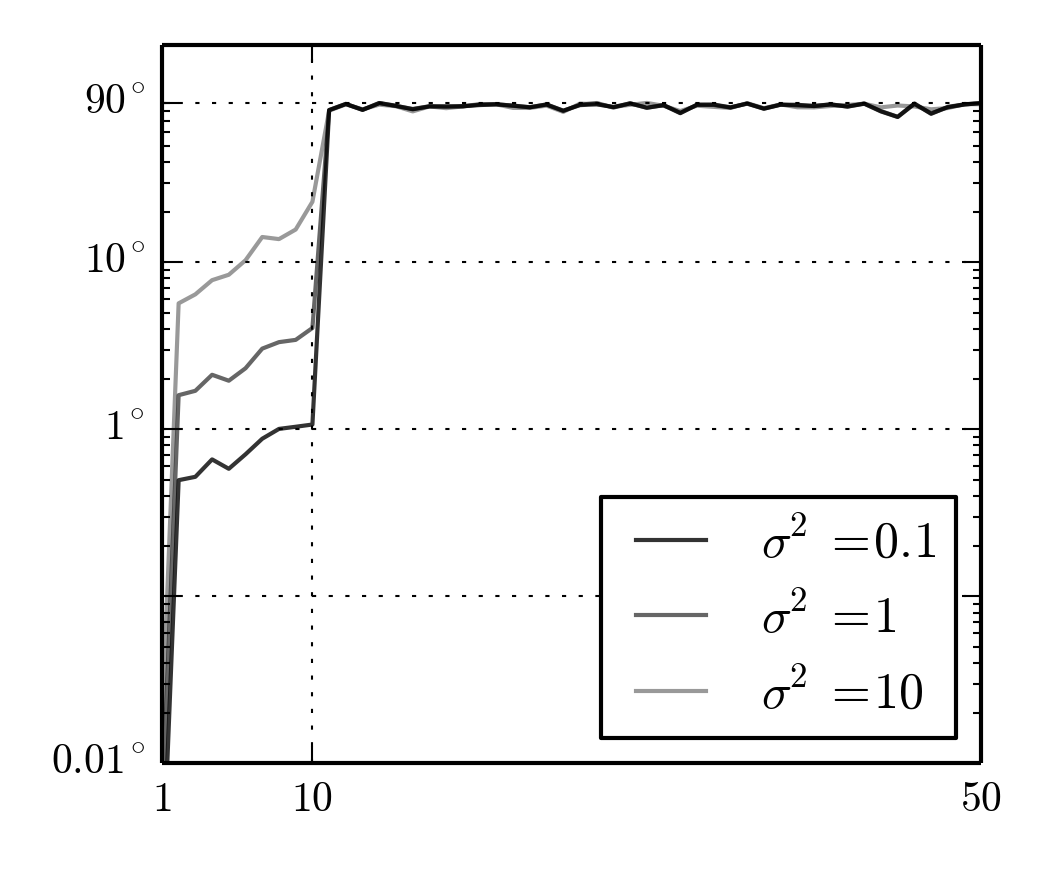}
    	\label{fig:signal_subspace:eigvectors}}
	\caption{The impact of noise on the EVD of the covariance matrix $\frac{1}{n} YY^\T$ using $m=10$.}
	\label{fig:signal_subspace}
\end{figure}

\autoref{fig:signal_subspace} visualizes this for $n = 10{,}000$ artificial observations, randomly generated accordingly to Eq.~\eqref{eq:model} using $m = 10$, $\shape a = 70$, $\shape y = 1000$, and a ground truth vector $x_\text{true}$ with values distributed uniformly between $0$ and $255$. In \autoref{fig:signal_subspace:eigvalues}, the eigenvalues $\lambda_i \geq \lambda_{i+1}$ of the covariance matrix \inline{$\frac{1}{n} YY^\T$} are plotted for different noise levels. The kink at the $10$-th eigenvalue marks the signal subspace dimension $m$. It can be seen that the kink is less clear for higher noise levels. The large drop-off after the first eigenvalue in all four curves in Figure~\ref{fig:signal_subspace:eigvalues} occurs because the vectors $y_i$ are not zero-mean. The eigenvector, which corresponds to the first eigenvalue, points roughly towards the mean of the observations. Since all other eigenvectors must be orthogonal, the corresponding eigenvalues must be of lower magnitude. \autoref{fig:signal_subspace:eigvectors} shows that the misalignments of the eigenvectors is larger for higher noise levels. It can also be seen that eigenvectors, which correspond to greater eigenvalues, tend to be more \emph{reliable} (i.e.\ less affected by noise).

\subsubsection{Application to image data}
\label{chap:signal_subspace:imagedata}

\begin{figure*} % exp094
	\centering
	\subfloat[$y_1$]{
		\includegraphics[width=0.181\textwidth]{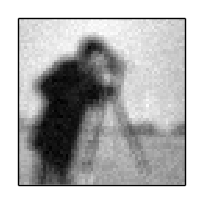}
    	\label{fig:eigfaces:y01}}
	\subfloat[$u_1$]{
		\includegraphics[width=0.181\textwidth]{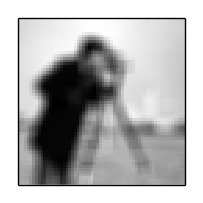}
    	\label{fig:eigfaces:u01}}
	\subfloat[$u_2$]{
		\includegraphics[width=0.181\textwidth]{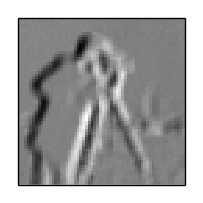}
    	\label{fig:eigfaces:u02}}
    \raisebox{12mm}[\height][\depth]{\enspace\dots}
	\subfloat[$u_{25}$]{
		\includegraphics[width=0.181\textwidth]{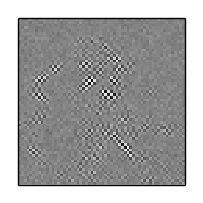}
    	\label{fig:eigfaces:u25}}
	\subfloat[$u_{26}$]{
		\includegraphics[width=0.181\textwidth]{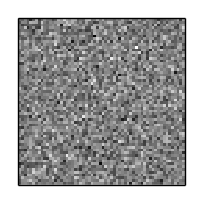}
    	\label{fig:eigfaces:u26}}
	\caption{An exemplary observation and the corresponding eigenvectors $u_1, \dots, u_{26}$ of the covariance matrix $\frac{1}{n} YY^\T$, where the observations $y_1, \dots y_{2500}$ were generated according to Eq.~\eqref{eq:model} using random $5 \times 5$ PSFs with $m = 25$. The pixels in the panels \ref{fig:eigfaces:y01} and \ref{fig:eigfaces:u01} are non-negative. In the panels \ref{fig:eigfaces:u02}--\ref{fig:eigfaces:u26}, $0$-valued pixels are colored grey, black corresponds to negative values, white to positive.}
	\label{fig:eigfaces}
\end{figure*}

\autoref{fig:eigfaces} illustrates the characteristics of the signal subspace described above for synthetic image data comprising $n = 2500$ noisy observations. The observations $y_1, \dots, y_n$ were created from the ground truth image in \autoref{fig:intro:cameraman}, sampled down to $\shape x = 64 \times 64$, and using randomly generated PSFs with $m = 25$. An exemplary observation is shown in \autoref{fig:eigfaces:y01}. The eigenvector $u_1$ in \autoref{fig:eigfaces:u01}, which corresponds to the greatest eigenvalue, is very blurry and noise-free like the mean of all observations. \autoref{fig:eigfaces:u26} confirms that the eigenvectors from $u_{m+1 = 26}$ on do not contain much information regarding the signal subspace but mostly noise.

\section{Likelihood maximization approach}
\label{chap:approach}

In this section, we describe our direct approach for MFBD using a non-convex loss function and without estimating the filters. We first describe the estimate $\hat x_{\mathsf n}$ of the ground truth $x_\text{true}$, which explains the observations $Y = \left[y_1\, \dots\, y_n\right]$ best in terms of likelihood maximization. The subscript $\mathsf n$ indicates, that $\hat x_{\mathsf n}$ aims to explain the original, noisy input data. Afterwards, we will use the results from \autoref{chap:signal_subspace:noisy}--\ref{chap:signal_subspace:imagedata} to derive the estimate $\hat x$ for the noise-free observations using the noisy image data.

Consider the probability $p\left(Y\middle|x, a_1,\dots,a_n\right)$ of the observations, given an undeteriorated image $x$ and the filters $a_1, \dots, a_n$. We confine ourselves to those parameters, for which the condition $p\left(Y\middle|x, a_1,\dots,a_n\right) \neq 0$ holds. Using the monotonicity of the logarithm, we then define
\begin{gather}
	\hat x_{\mathsf n} =
		\arg\min_x \min_{a_1,\dots,a_n} -\ln p\left(Y\middle|x, a_1,\dots,a_n\right)
	\text.
    \label{eq:ml_log}
\end{gather}
Strictly speaking, the minimizer of Eq.~\eqref{eq:ml_log} is not necessarily unique (cf. \autoref{chap:approach:solution_ambiguity}), so $\hat x_{\mathsf n}$ is simply defined as an arbitrary minimizer. Assuming that the observations $y_1, \dots, y_n$ are statistically independent, they factorize like $p\left(Y\middle|x, a_1,\dots,a_n\right) = \prod_{i=1}^n p\left(y_i\middle|x, a_1,\dots,a_n\right)$, and hence
\begin{gather}
	\hspace*{-1mm}
	\hat x_{\mathsf n}
	=	\arg\min_x \min_{a_1,\dots,a_n} -\sum_{i=1}^n \ln p\left(y_i\middle|x, a_1,\dots,a_n\right)
	\text.
	\label{eq:ml_factorized}
\end{gather}
is obtained. In \autoref{chap:model}, we assumed that $\varepsilon_i$ is AWGN, so Eq.~\eqref{eq:model} takes the form $y_i \sim \mathcal N\left(a_i \ast x; I\sigma^2\right)$. Plugging this into Eq.~\eqref{eq:ml_factorized} in place of $p\left(y_i\middle|x, a_1,\dots,a_n\right)$ and dropping those terms from the objective function, which are constant w.r.t.\ $x$ and $a_i$, the likelihood maximization approach for $\hat x_{\mathsf n}$ boils down to the least-squares problem
\begin{gather}
    \hat x_{\mathsf n}
	=	\arg\min_x \min_{a_1,\dots,a_n} \sum_{i=1}^n \left\|y_i - a_i \ast x\right\|^2
	\label{eq:lsq_approach}
	\text.
\end{gather}

So far, the approach is canonical and similar to the work from \citet{harikumar1999} and \citet{harmeling2009}. Eq.~\eqref{eq:lsq_approach} requires the joint optimization w.r.t.\ $a_1, \dots a_n$ and $x$. We will simplify this problem by confining the parameter space to such $a_1 \dots a_n$, which fulfill the necessary condition for the presence of a minimum in $x$, as described below.

\subsection{Closed-form constraint for $a_i$}

We derive the differential $\differential \left\|y_i - a_i \ast x\right\|^2 = \differential (\left(y_i - \cnvmat Xa_i\right)^\T \left(y_i - \cnvmat Xa_i\right))$ of the summands of the objective function in Eq.~\eqref{eq:lsq_approach} using \cite{magnus1999}, that is
\begin{gather}
	\differential \left\|y_i - a_i \ast x\right\|^2 = -2 \left(y_i - \cnvmat Xa_i\right)^\T \differential a_i
	\text.
	\label{eq:lsq_differential_ai}
\end{gather}
From Eq.~\eqref{eq:lsq_differential_ai}, we can read off the derivative of $\left\|y_i - a_i \ast x\right\|^2$ w.r.t.\ $a_i$ and use it to find the closed-form optimization constraint on $a_i$ for the presence of a minimum,
\begin{gather}
	\frac{\partial}{\partial a_i} \left\|y_i - a_i \ast x\right\|^2
	= -2 \left(y_i - \cnvmat Xa_i\right)^\T = 0
	\text,
\end{gather}
that is $y_i = \cnvmat X a_i$. The null space of $\cnvmat X^\T$ is orthogonal to the range of $\cnvmat X$. To see this, consider a vector $v$ from the null space of $\cnvmat X^\T$, i.e.\ $\cnvmat X^\T v = 0$. This means that $v$ is orthogonal to the range of $\cnvmat X$. Thus, and since $y_i \in \range \cnvmat X$, prepending $\cnvmat X^\T$ to both sides of the equation doesn't affect its solution for $a_i$:
\begin{gather}
	\frac{\partial}{\partial a_i} \left\|y_i - a_i \ast x\right\|^2 = 0
	\enspace\Leftrightarrow\enspace \cnvmat X^\T y_i = \cnvmat X^\T \cnvmat X a_i
\end{gather}
The assumption $\rank \cnvmat X = \shape a$ from \autoref{chap:signal_subspace} implies that $\cnvmat X^\T \cnvmat X$ has full rank \cite{luetkepohl1996}, so $\cnvmat X^\T \cnvmat X$ is invertible and its inverse has full rank too. Thus, prepending \inline{$\left(\cnvmat X^\T \cnvmat X\right)^{-1}$} to both sides yields another equivalent equation:
\begin{gather}
	\left(\cnvmat X^\T \cnvmat X\right)^{-1} \cnvmat X^\T y_i = a_i
	\label{eq:a_closed_form_solution}
\end{gather}
Plugging Eq.~\eqref{eq:a_closed_form_solution} back into the least squares form in Eq.~\eqref{eq:lsq_approach} yields an expression, which only needs to be minimized w.r.t.\ $\cnvmat X$. After resolving the squared norm $\left\|\cdot\right\|^2$ using the inner vector product and dropping those summands, which are constant w.r.t.\ $\cnvmat X$, we finally obtain the estimate
\begin{gather}
	\hat x_{\mathsf n}
	=	\arg\max_x \sum_{i=1}^n
		y_i^\T \cnvmat X \left(\cnvmat X^\T \cnvmat X\right)^{-1} \cnvmat X^\T y_i
		\label{eq:lsq_final}
	\text,
\end{gather}
where each $\cnvmat X$ depends linearly on $x$, as described in \autoref{chap:boundary_treatment}.

\subsection{Denoising}

The summands \inline{$y_i^\T \cnvmat X \left(\cnvmat X^\T \cnvmat X\right)^{-1} \cnvmat X^\T y_i$} in Eq.~\eqref{eq:lsq_final} are scalar-valued. Using the trace operator $\trace$, the estimate $\hat x_{\mathsf n}$ is stated equivalently as that $x$ which maximizes \inline{$\trace \sum_{i=1}^n y_i y_i^\T \cnvmat X \left(\cnvmat X^\T \cnvmat X\right)\!{}^{-1} \cnvmat X^\T$}. Using $YY^\T = \sum_{i=1}^n y_i y_i^\T$ yields
\begin{gather}
	\hat x_{\mathsf n}
	=	\arg\max_x
		\trace YY^\T \cnvmat X \left(\cnvmat X^\T \cnvmat X\right)^{-1} \cnvmat X^\T
	\text.
\end{gather}
We now replace $YY^\T$ by $U \Lambda U^\T$, that is its EVD, but truncate the $\shape y \times \shape y$ matrix $U$ after its $m$ first columns, and resolve the trace-operator. This yields the estimate
\begin{gather}
	\hat x =
		\arg\max_x \sum_{i=1}^{m}
		\lambda_i u_i^\T \cnvmat X \left(\cnvmat X^\T \cnvmat X\right)^{-1} \cnvmat X^\T u_i,
		\label{eq:approach_final}
\end{gather}
which, in view of \autoref{chap:signal_subspace} and for $n \to \infty$, maximizes the likelihood of the \emph{noise-free} observations. We will hence refer to $\hat x$ as the \emph{denoised} estimate.

\subsection{Geometric interpretation}

\begin{figure}[htb]
  \begin{center}
  \includegraphics[scale=0.66]{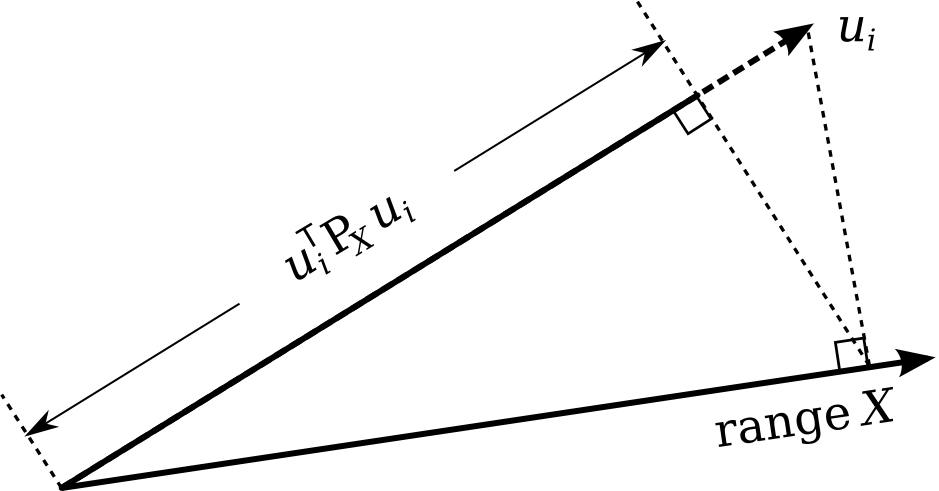}
  \caption{Geometric interpretation of the estimate $\hat x$ in Eq.~\eqref{eq:approach_final}. The dashed lines indicate projections according to Eq.~\eqref{eq:approach_geom_interp} with $\left\|\mathrm{P}_{\cnvmat X} u_i\right\|^2 = u_i^\T \mathrm{P}_{\cnvmat X} u_i$ for the simplified case $\rank \cnvmat X = 1$.}
  \label{fig:approach:geom_interp}
  \end{center}
\end{figure}

Note that \inline{$\mathrm{P}_{\cnvmat X} = \cnvmat X \left(\cnvmat X^\T \cnvmat X\right)^{-1} \cnvmat X^\T$} is the \emph{projector} onto the range of $\cnvmat X$, if $\rank \cnvmat X = \shape a$ (e.g., \cite{luetkepohl1996}), which leads us to
\begin{gather}
	\hat x = \arg\max_x \sum_{i=1}^m \lambda_i u_i^\T \mathrm{P}_{\cnvmat X} u_i
	\label{eq:approach_geom_interp_quad_form}
	\text.
\end{gather}
The matrix $\cnvmat X^\T \cnvmat X$ is symmetric. It is easily seen that its inverse, and consequently also $\mathrm{P}_{\cnvmat X}$, are symmetric too. Since $\mathrm{P}_{\cnvmat X}$ is idempotent, i.e. $\mathrm{P}_{\cnvmat X}^2 = \mathrm{P}_{\cnvmat X}$, we get $\hat x = \arg\max_x \sum_{i=1}^m \lambda_i u_i^\T \mathrm{P}_{\cnvmat X}^\T \mathrm{P}_{\cnvmat X} u_i$, which we rewrite as
\begin{gather}
	\hat x
    =	\arg\max_x \sum_{i=1}^m \lambda_i \left\|\mathrm{P}_{\cnvmat X} u_i\right\|^2
    \text.
    \label{eq:approach_geom_interp}
\end{gather}
Eq.~\eqref{eq:approach_geom_interp} dictates, that the longer the projections of the eigenvectors $u_i$ onto the range of $\cnvmat X\!\left(x\right)$ are, the better $x$ explains the observations. The term \inline{$\left\|\mathrm{P}_{\cnvmat X} u_i\right\|\!{}^2$} indicates how good $x$ explains variance along $u_i$, as \autoref{fig:approach:geom_interp} illustrates for the simplified case $\rank \cnvmat X = 1$. In view of \autoref{chap:signal_subspace}, the factors $\lambda_i$ induce the following weighting: A good explanation for larger variance outweighs an equally good explanation for smaller variance.

\subsection{Solution ambiguity}
\label{chap:approach:solution_ambiguity}

We also see from Eq.~\eqref{eq:approach_geom_interp} that, in general, there is an infinite number of solutions $x$: The objective function is invariant w.r.t.\ the multiplication of $x$ by a scalar $\alpha$ (i.e.\ the brightness of $\hat x$ is ambiguous). This arises from the homogeneity $\left(\cnvmat X \alpha\right) a_i = \cnvmat X \left( \alpha a_i \right)$ of convolution in the underlying model. %\textcolor{red}{We will tackle this ambiguity by choosing the brightness of $\hat x$ equally to the observations' mean.}

\subsection{Gradient ascent}
\label{chap:approach_direct_solution}

This section demonstrates, that the direct solution of Equation~\eqref{eq:approach_final} is difficult, due to the presence of local extrema in the objective function. To simplify notation, we write \inline{$\phi\left(x\right) = \sum_{i=1}^n \lambda_i \phi_i\left(x\right)$} and \inline{$\phi_i\left(x\right) = u_i^\T \cnvmat X \left(\cnvmat X^\T \cnvmat X\right)\!{}^{-1} \cnvmat X^\T u_i$} to refer to the objective function in Eq.~\eqref{eq:approach_final} and its summands.

Using \cite{magnus1999} and Eq.~\eqref{eq:convolution:valid} to resolve the matrix $\cnvmat X$, we obtain the differential
\begin{gather}
	\differential \phi_i\left(x\right)
	=	2 r_i^\T I_y^\T F^\Hermitian \Diag \left( F I_a v_i \right) F \differential x
	\text,
	\label{eq:approach_gradient_wrt_x}
\end{gather}
where we abbreviate \inline{$v_i = \left(\cnvmat X^\T \cnvmat X\right)^{-1} \cnvmat X^\T u_i$} and \inline{$r_i = u_i - \cnvmat Xv_i$}. Then, we can read off the gradient from Eq.~\eqref{eq:approach_gradient_wrt_x},
\begin{gather}
	\nabla \phi_i\left(x\right)
	=	\frac{\partial}{\partial x} \phi_i\left(x\right)
	=	2 r_i^\T I_y^\T F^\Hermitian \Diag \left( F I_a v_i \right) F
	\text.
\end{gather}
To avoid the costly computation of the DFT matrix $F$, we rewrite $\nabla \phi_i\left(x\right)$ as \inline{$\left(2 F^\T \Diag \left( F I_a v_i \right) \conj{F} I_y r_i\right)^\T$}. Putting $r_i = u_i - \cnvmat Xv_i$ back in and using that $\phi_i = \conj{\phi_i}$ since $\phi_i$ is real-valued yields
\begin{gather}
	\nabla \phi_i\left(x\right)
	=	\left(2 F^\Hermitian \Diag \left( \conj{F I_a v_i} \right) F I_y \left(u_i - \cnvmat Xv_i\right)\right)^\T
	\text.
	\label{eq:approach_gradient_wrt_x_final}
\end{gather}

\begin{figure*}
    \centering
	\includegraphics[height=35mm, trim=0 220px 0 0, clip]{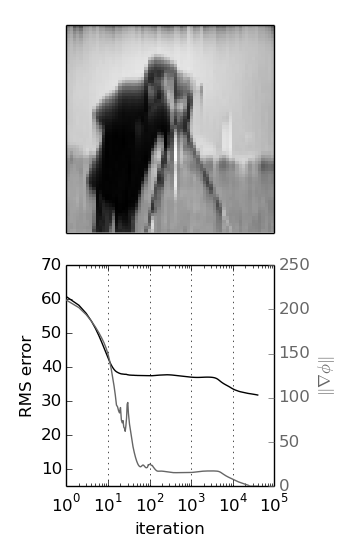}
	\includegraphics[height=35mm, trim=0 0 0 170px, clip]{bilder/grad_ascent_on_u}
	\caption{Gradient ascent performance using a randomly generated image sequence ($n = 1000$ synthetic observations). The image (left) corresponds to the final estimate $\hat x_{t}$ after $t = 10^5$ iterations. The plot (right) shows the RMS error curve of the estimate $\hat x_t$ as well as the norm of the gradient $\nabla\phi$.}
	\label{fig:grad_ascent}
\end{figure*}

The gradient of the objective function $\phi\left(x\right)$ always points into the direction of the steepest ascent. Given an estimate $\hat x_0$, the \emph{gradient ascent} iteration \inline{$\hat x_{t+1} = \hat x_t + \tau \nabla \phi\left(\hat x_t\right)$} yields an improved estimate. The parameter $\tau$ controls the step distance per iteration, we used $\tau = 1$ (for details, see e.g.\ \cite{murphy2012}). \autoref{fig:grad_ascent} shows the result after $10^5$ iterations using a randomly generated initialization $\hat x_0$. The corresponding error and $\left\|\nabla \phi\right\|$ curves indicate, that a non-global peak is reached after around $2 \cdot 10^4$ iterations. Such non-global extrema hamper the search for the global maximum of $\phi$, unless a good initialization is known a priori.

\section{Eigenvector method}
\label{chap:method}

So far, we have derived an optimization problem based on likelihood maximization, whose solution $\hat x$ recovers the undeteriorated image from its blurry and noisy observations. We have applied an iterative ascending method and found that the computation of $\hat x$, as we have formulated it so far, is difficult, due to the presence of local extrema in the objective function $\phi$. In this section, we study a different objective function, which is computationally easier to optimize. We will see, that -- under specific conditions -- this optimization problem is equivalent to the likelihood maximization-based method described in \autoref{chap:approach}.

We start by rewriting Eq.~\eqref{eq:approach_geom_interp_quad_form} as the minimization of $-\sum_{i=1}^m \lambda_i u_i^\T \mathrm P_{\cnvmat X} u_i - \lambda_i u_i^\T u_i$, where the summands $-\lambda_i u_i^\T u_i$ may be added to the objective function without affecting the solution since they are constant w.r.t.\ $x$. This yields $\hat x = \arg\min_x \sum_{i=1}^m \lambda_i u_i^\T \left(I - \mathrm P_{\cnvmat X}\right) u_i$, where we recognize $\mathrm P_{\cnvmat X^\perp} = I - \mathrm P_{\cnvmat X}$ as the projector onto the orthogonal complement of the range of $\cnvmat X$ (see, e.g., \cite{luetkepohl1996}). Using the idempotency and symmetry of $\mathrm P_{\cnvmat X^\perp}$, we obtain
\begin{gather}
	\hat x = \arg\min_x \varphi\left(x\right),
    \qquad
    \varphi\left(x\right) = \sum_{i=1}^m \lambda_i \left\|\mathrm{P}_{\cnvmat X^\perp} u_i\right\|^2
	\label{eq:approach_as_minimization}
	\text.
\end{gather}
In the following, we will consider three cases, which will be explained in more detail, when they come into play:
\begin{enumerate}
	\item We start with the idealistic, noise-free case $m = \shape a$.
	\item We consider the still idealistic, noisy case $m = \shape a$.
	\item Finally, we study the realistic case $m < \shape a$.
\end{enumerate}
The case $m > \shape a$ can only occur, if either $\shape a$ is underestimated or $m$ is determined incorrectly. Then, revisiting $m$ or $\shape a$ yields one of the other cases.

\subsection{Noise-free $m = \shape a$ case}
\label{chap:method:noise_free}

Recall from \autoref{chap:theory} that in the noise-free case, any observation $y_i = Xa_i$ is a linear combination of the columns of the matrix $X = X\!\left(x_\text{true}\right)$, and thus $\range U \subseteq \range \cnvmat X$. This means that the minimum of the objective function $\varphi$ in Eq.~\eqref{eq:approach_as_minimization} is $\varphi\left(\hat x\right) = 0$. Since $m$ was defined in \autoref{chap:signal_subspace} so that $\lambda_i > 0 $ for all $i \leq m$, it is seen that $\varphi\left(x\right) = 0$ can occur if and \emph{only} if $\mathrm P_{\cnvmat X^\perp} u_i = 0$ for all $i \leq m$. Thus, $\varphi\left(x\right) = 0$ occurs not just if, but also \emph{only} if $\range U \subseteq \range \cnvmat X$.

Consider the statement, that $\mathrm P_{U^\perp} x_k = 0$ for all $k$, where the vectors $x_1, \dots, x_{\shape a}$ are the columns of the matrix $\cnvmat X = \cnvmat X\!\left(x\right)$ and thus span the range of $\cnvmat X$. Clearly, this statement is true if and only if $x$ is chosen so that $\range \cnvmat X \subseteq \range U$. The inclusion ``$\subseteq$'' tightens to the equality $\range U = \range \cnvmat X$ if $x = x_\text{true}$ and $m = \shape a$, as it is seen from Eq.~\eqref{eq:model}. Thus, for $m = \shape a$ and in the absence of noise, we obtain the fundamental equivalence
\begin{gather}
	\varphi\left(x\right) = 0
	\enspace\Leftrightarrow\enspace
	\mathrm P_{U^\perp} x_k = 0 \enspace\forall k
	\text.
	\label{eq:method_solution_equivalence}
\end{gather}
The eigenvalues $\lambda_1, \dots, \lambda_m$ only appear in $\varphi$, but not on the right-hand side of the ``$\Leftrightarrow$''.

The fundamental equivalence~\eqref{eq:method_solution_equivalence} means that, for $m = \shape a$ and in the absence of noise, we can solve the original minimization problem~\eqref{eq:approach_as_minimization} by instead minimizing the residuals \inline{$\left\| \mathrm P_{U^\perp} x_k \right\|\!{}^2$}, i.e.
\begin{gather}
    \hat x
    =	\arg\min_x
    	\sum_{k=1}^{\shape a} \left\| \mathrm P_{U^\perp} x_k \right\|^2
    		\enspace\text{s.t.}\enspace \hat x^\T \hat x = 1
    \text,
    \label{eq:method_abstract}
\end{gather}
where we add the constraint $\hat x^\T \hat x = \left\|\hat x\right\| = 1$ to avoid the trivial solution $\hat x=0$. We are allowed to do this, because the optimization problems from Eq.~\eqref{eq:method_abstract} and~\eqref{eq:approach_as_minimization} are solved by the same $x$, and as pointed out in \autoref{chap:approach:solution_ambiguity}, the value of the objective function is invariant to scalar factors.

To construct the vectors $x_k$, we define an $\shape y \times \shape x$ matrix $\cnvmat B_k$ so that \inline{$\cnvmat B_k x = b_k \ast_\text{valid} x$}, where $b_k \in \mathbb R^{\shape a}$ is the $k$-shifted Kronecker delta, i.e.\ with $b_k\!\left[i\right] = \left\{\text{$1$ if $k=i$; $0$ else}\right\}$ (see, e.g., \cite{kamisetty2001}). Then, the linearity of convolution implies that $\SPAN \left\{\cnvmat B_k x\middle|k\right\} = \SPAN \left\{\cnvmat A_i x\middle|i\right\}$ for $m = \shape a$. From the commutativity of convolution $\cnvmat A_i x = \cnvmat X a_i$ we see in particular, that $x_k = \cnvmat B_k x$, and obtain
\begin{gather}
    \hat x
    =	\arg\min_{x}
    	x^\T M\, x \enspace\text{s.t.}\enspace \hat x^\T \hat x = 1,
    \qquad
    M = \sum_{k=1}^{\shape a} \cnvmat B_k^\T \left(I - UU^\T\right) \cnvmat B_k
    \text.
    \label{eq:method_final}
\end{gather}
Due to the constraint $\hat x^\T \hat x = 1$, the objective function in Eq.~\eqref{eq:method_final} is recognized as the \emph{Rayleigh quotient} \inline{$x^\T M\, x / x^\T x$}. Since the matrix $M$ is symmetric, the \emph{Rayleigh-Ritz theorem} \cite{luetkepohl1996} states that $\hat x$ is the eigenvector of $M$, which corresponds to its smallest eigenvalue. This eigenvalue is the value of the objective function for $x = \hat x$, that is $\varphi\left(\hat x \right) = 0$. Since the matrix $M$ is positive semidefinite due to Eq.~\eqref{eq:method_abstract}, all its eigenvalues are real and non-negative. There may be other eigenvectors for eigenvalue $0$, but not if the original optimization problem's solution is unique up to a scalar factor.

\subsection{Noisy $m = \shape a$ case}
\label{chap:method:noisy_ideal}

In the previous section, we have seen that the objective function $\sum_{k=1}^{\shape a} \left\| \mathrm P_{U^\perp} x_k \right\|^2$ of Eq.~\eqref{eq:method_abstract} is $0$, if and only if all columns $x_1, \dots, x_{\shape a}$ of the matrix $\cnvmat X = \cnvmat X\!\left(x_\text{true}\right)$, can be represented as linear combinations of the eigenvectors $u_1, \dots, u_m$. We express such a linear combination as $x_k = U c_k$ with a weighting vector $c_k$, where $c_k\!\left[i\right]$ corresponds to the \emph{contribution} of $u_i$ to $x_k$. %The equivalence \eqref{eq:method_solution_equivalence} shows that such representation exists for $x = x_\text{true}$.

It was shown in \autoref{chap:signal_subspace} that for higher noise levels (and a small number $n$ of images), the signal subspace is reflected less truthfully and the eigenvectors encoded in the matrix $U$ become misaligned. Although this error can be kept small by increasing the number $n$ of observations, in practical applications, at least a small error $\rho_i$ always remains on every eigenvector $u_i$, since the number of observations must be finite. Looking close at the eigenvectors shown in \autoref{fig:eigfaces:u01}--\ref{fig:eigfaces:u25}, one can see that this error appears as ``noisy'' grain. We will quantify the error as $\rho_i \sim \mathcal N\left(0; \Sigma_i\right)$ without assuming that $\rho_i$ is i.i.d., because this would disrespect the orthonormal nature of the eigenvectors $u_1, \dots, u_m$.

In general, a vector $c_k$ which satisfies $x_k = U c_k$ for $x = x_\text{true}$ does not necessarily exists for all $k$ when the columns of $U$ are misaligned due to noise. Still, it does exist for $x_k = \left[u_1 - \rho_1\, \dots\, u_m - \rho_m\right] c_k$, where $u_i - \rho_i$ are the unknown, error-free eigenvectors. By rewriting $\left[u_1 - \rho_1\, \dots\, u_m - \rho_m\right] c_k$ as $U c_k - \sum_{i=1}^m c_k\!\left[i\right] \rho_i$, we see that
\begin{gather}
	x_k = U c_k - \delta_k
	\text,
	\label{eq:method_noisy_residual}
\end{gather}
where $\delta_k = \sum_{i=1}^m c_k\!\left[i\right] \rho_i$ is the normal-distributed, zero-mean residual with
\begin{gather}
	\Cov \delta_k
	=	\sum_{i=1}^m \Sigma_i c_k\!\left[i\right]^2
	\text.
	\label{eq:method_noisy_residual_cov}
\end{gather}

If the errors $\rho_1, \dots, \rho_m$ of the eigenvectors $u_1, \dots, u_m$ do not occur to be linear combinations of the eigenvectors, i.e.\ $\rho_i \not\in \range U$ for all $i = 1, \dots, m$, then we get $\delta_k \not\in \range U$, since $\delta_k$ is a linear combination of the errors. We also get $\delta_k \not\in \range U$ if only some errors $\rho_i$ are not in the range of $U$, as long as these $\rho_i$ are not zero-weighted by $c_k$. According to Eq.~\eqref{eq:method_noisy_residual}, we can write the objective function as \inline{$\sum_{k=1}^{\shape a} \left\| \mathrm P_{U^\perp} x_k \right\|\!{}^2 = \sum_{k=1}^{\shape a} \left\| \mathrm P_{U^\perp} \delta_k \right\|\!{}^2$}, because $\mathrm P_{U^\perp} U c_k = 0$. Consequently, when the noise level rises and the eigenvector errors $\rho_i$ grow, the value of the objective function for $x = x_\text{true}$ becomes greater than $0$.

Since choosing $x = \hat x$ minimizes $\sum_{k=1}^{\shape a} \left\| \mathrm P_{U^\perp} x_k \right\|^2 \eqrefrel{eq:method_noisy_residual}{=} \sum_{k=1}^{\shape a} \left\| \mathrm P_{U^\perp} \delta_k \right\|^2$ by definition, the covariances $\Cov \delta_k$ of the residuals are also minimized. Due to Eq.~\eqref{eq:method_noisy_residual_cov}, this induces a preference for those $c_k$ which assign a small weight $c_k\!\left[i\right]$ for $\Sigma_i$ if $\Sigma_i$ is large. This means that $c_k$ tends to weight the columns of $U$ in accordance to their \emph{reliability}, and the most reliable eigenvector is $u_1$ (see \autoref{chap:signal_subspace:noisy}). Therefore, when the noise level is increased, the \emph{contribution} from $u_1$ tends to be overestimated -- which makes $\hat x$ become more blurry, but not noisier than $u_1$. Surprisingly, this mechanism works without taking the eigenvalues $\lambda_1, \dots, \lambda_m$ into account, which encode the reliability of the corresponding $u_i$. Inventing a mechanism to counter-balance the overestimation remains an open problem for future research.

\subsection{Noisy $m < \shape a$ case}
\label{chap:method:noisy_real}

So far, we have confined ourselves to the case $m = \shape a$ which yields the equality $\range U = \range \cnvmat X$ for $X = \cnvmat X\!\left(x_\text{true}\right)$. We have seen, how this can be utilized to minimize \inline{$\sum_{k=1}^{\shape a} \left\| \mathrm P_{U^\perp} x_k \right\|\!{}^2$} instead of \inline{$\sum_{i=1}^m \left\| \mathrm P_{\cnvmat X^\perp} u_i \right\|\!{}^2$}. But as we already mentioned in Section~\ref{chap:model}, the assumption $m = \shape a$ is rarely true in practice, so the strict inclusion $\range U \subset \range \cnvmat X$ is a rather realistic condition for $X = \cnvmat X\!\left(x_\text{true}\right)$. The fundamental equivalence \eqref{eq:method_solution_equivalence} does not hold then.

The case $m < \shape a$ can always be seen as the situation that an insufficient amount of observations was acquired. As we described in \autoref{chap:signal_subspace}, the signal subspace dimension $m$ equals the dimension of the PSF subspace $\SPAN \left\{a_1, \dots, a_n\right\}$ if $x_\text{true}$ is persistently exciting. Thus, encountering $m < \shape a$ means that the observations were generated from PSFs which did not contain enough variance. Acquiring additional observations generated from the ``missing'' PSFs would establish the $m = \shape a$ case. To some extent, such acquisition can be synthesized, as described below.

\subsubsection{Inflating the observations}
\label{chap:method:inflating}

The rule of associativity in Eq.~\eqref{eq:convolution:associativity} allows us to artificially generate additional observations with unobserved PSFs. To accomplish this, we transform each observation $y_i$ using a valid convolution matrix $\cnvmat D_t$, whose underlying filter is a $t$-shifted Kronecker delta, i.e.\ with $d_t\!\left[i\right] = \left\{\text{$1$ if $t=i$; $0$ else}\right\}$ of size $\shape d$ (see, e.g., \cite{kamisetty2001}). According to Eq.~\eqref{eq:model},
\begin{gather}
	\cnvmat D_t y_i \eqrefrel{eq:convolution:associativity}{=} \left(d_t \ast_\text{full} a_i\right) \ast_\text{valid} x_\text{true} + \cnvmat D_t \varepsilon_i
	\label{eq:inflating}
\end{gather}
shifts the PSF $a_i$, which generated the observation $y_i$, by the offset of the Kronecker delta. By \emph{inflating} we mean the substitution of the original observations by $\cnvmat D_1 Y, \dots, \cnvmat D_{\shape d} Y$. However, inflating not only increases the signal subspace dimension $m$, but, due to full convolution, also increases the dimension of the PSFs (see \autoref{chap:boundary_treatment}). Below, $m$, $\shape a$, and $\shape y$ refers to the respective quantities after inflating, and, to avoid confusion, we will write $m_0$, $\shape{a,0}$, and $\shape{y,0}$ to refer to the original quantities.

\begin{figure}
	\centering
	\subfloat[$m_0=5$]{
    	\includegraphics[scale=0.5, trim=0 25px 0 0, clip]{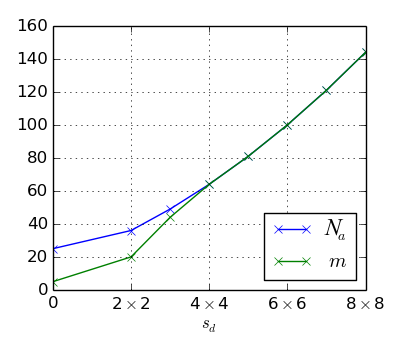}
		\label{fig:reblur:random_submanifolds:m5_noise0.1}}
	\subfloat[$m_0=10$]{
    	\includegraphics[scale=0.5, trim=0 25px 0 0, clip]{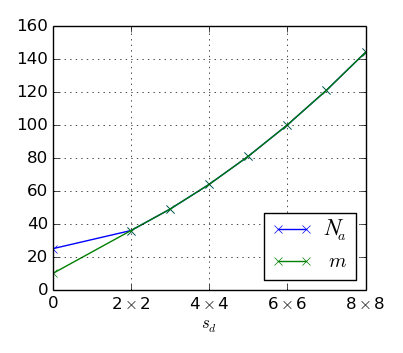}
		\label{fig:reblur:random_submanifolds:m10_noise0.1}}
	\caption{The plots show how the PSF dimension $\shape a$ and the signal subspace dimension $m$ typically grow, when each observation $y_i$ is inflated according to Eq.~\eqref{eq:inflating} using different sizes $\shape d$ of the inflating filter $d_t$ (horizontal axes). The original PSFs were $\shape{a,0} = 5 \times 5$ pixels in size and sampled from a random, $m_0$-dimensional subspace of $\mathbb R^{\shape{a,0}}$. The plots were computed using $n=10{,}000$ noisy observations of the ground truth in \autoref{fig:intro:cameraman}, sampled down to $\shape x = 32 \times 32$ pixels, with $\varepsilon_i \sim \mathcal N\left(0; I \cdot 10^{-1}\right)$.}
	\label{fig:reblur:random_submanifolds}
\end{figure}

\autoref{fig:reblur:random_submanifolds} shows the typical behavior of $\shape a$ and $m$ after inflating in dependence of the filter size $\shape d$ for randomly generated, $m_0$-dimensional PSFs with $m_0 < \shape{a,0}$. The signal subspace dimension $m$ was estimated as the rank of the matrix $Y$. As can be seen from comparison of \autoref{fig:reblur:random_submanifolds:m5_noise0.1} and \autoref{fig:reblur:random_submanifolds:m10_noise0.1}, an initially greater signal subspace dimension $m_0$ facilitates, that smaller filter sizes $\shape d$ suffice for reaching the desired $m = \shape a$ state (for fixed $\shape{a,0}$).

\subsubsection{When inflating is not enough}
\label{chap:when_inflating_fails}

Depending on the PSF subspace, it might be that the $m = \shape{a}$ state cannot be reached by inflating (e.g., for PSFs with disk-shaped support). Generally, the more pixels at the corners of all observed PSFs are constant, the greater the gap of $m$ and $\shape a$ will remain after inflating. This can be tackled by reducing $\shape a$ in the one-dimensional case.

For images, we leave out those columns $x_k$ of the matrix $\cnvmat X$, which do not appear in the range of $U$, so that that the fundamental equivalence \eqref{eq:method_solution_equivalence} holds for $\cnvmat X = \cnvmat X\!\left(x_\text{true}\right)$. To do this, we introduce the matrix
\begin{gather}
	M_\ast = Q\!\left(h\right) + \alpha \cdot \sum_{k=1}^{\shape a} h\!\left[k\right] \cdot M_k,
    \qquad\text{where }
	M_k = \cnvmat B_k^\T \left(I - UU^\T\right) \cnvmat B_k
	\text,
	\label{eq:matrix_of_interest_generalized}
\end{gather}
which generalizes the matrix $M$ from \autoref{chap:method:noise_free}. Let $Q\!\left(h\right) = 0$ and $\alpha = 1$ for simplicity first. If the $\shape a$-shaped vector $h$, called the \emph{PSF footprint}, suffices
\begin{gather}
	h\left[k\right] =
	\begin{cases}
		0 & \text{if } a_i\left[k\right] = 0 \text{ for all } i, \\
		1 & \text{else,}
	\end{cases}
	\label{eq:psf_mask}
\end{gather}
then $M_\ast\, x_\text{true} \xrightarrow{n \to \infty} 0$. Yet, the minimization of $x^\T M_\ast\, x$ w.r.t.\ $x$ does not necessarily recover $x_\text{true}$, as the following example illustrates: Given that $a_i\left[1\right] = 0$ for all $i$ and $h = \left(0, 1, \dots, 1\right)$, the vector $\delta_{\shape x} = \left(0, \dots, 0, 1\right)$ also yields $M_\ast\, \delta_{\shape x} = 0$. This is because no information about the last pixel was observed.

Fortunately, the knowledge of the PSF footprint $h$ makes the solution of this ambiguity straight-forward, by choosing
\begin{gather}
    Q\!\left(h\right) = \Diag q\!\left(h\right),
    \qquad\text{where }
	q\!\left(h\right) = 1_{\shape x} - \min\left\{1_{\shape x},\ h \ast_\text{full} 1_{\shape y} \right\}
	\text.
	\label{eq:penalty_mask}
\end{gather}
Then, $x^\T Q\, x$ produces high responses for $x$ with non-zero values in those pixels, for which no information was observed. The minimization of $x^\T M_\ast\, x$ w.r.t.\ $x$ in accordance with Eq.~\eqref{eq:matrix_of_interest_generalized} and~\eqref{eq:penalty_mask} forces these pixels to $0$, and if $h$ and $U$ are accurate so that $m = \left\langle 1, h \right\rangle$, then the minimization
\begin{gather}
	\hat x_\ast = \arg\min_x x^\T M_\ast\, x
	\enspace\text{s.t.}\enspace
	x^\T x = 1
	\label{eq:generalized_estimate}
\end{gather}
yields the \emph{generalized} estimate $\hat x_\ast$ which recovers the ground truth $x_\text{true}$. Note that the condition $m = \left\langle 1, h \right\rangle$ is a generalization of $m = \shape a$ to the case that the PSF footprint $h$ may have $0$-entries, where $\left\langle\cdot,\cdot\right\rangle$ denotes the inner product.

\subsubsection{Estimating the PSF footprint}
\label{chap:estimating_psf_masks}

So far, we have described a solution of the MFBD problem based on the computation of the eigenvector of a specifically constructed matrix $M_\ast$. The only ingredient of this matrix, which remains unspecified, is the PSF footprint $h$. We propose determining the footprint $h$ heuristically by alternating minimization of $\hat x^\T M_k\, \hat x$ w.r.t.\ $\hat x$ and $h$. Note that this is different from alternating optimization w.r.t.\ the undeteriorated image $x$ and the filters $a_1, \dots, a_n$, since only one footprint needs to be determined for the whole image sequence.

The procedure is outlined in \autoref{alg:estimating_psf_masks}, which only takes the model parameter $\shape a$ and the truncated matrix of eigenvectors $U$ of the covariance matrix of the observations as input. The initialization of $\hat x_\ast$ using a rough estimate and the iterative refinement are described in \autoref{chap:method:impl:rayleigh} and \autoref{chap:method:impl:poweritr} below.

\begin{algorithm}
 	\caption{Outline of the computation of the generalized estimate $\hat x_\ast$ from an unknown PSF footprint $h$.}
	\label{alg:estimating_psf_masks}
 	
	\SetAlgoLined
	\KwIn{$\shape a$ and the truncated matrix of eigenvectors $U = \left(u_1, \dots, u_m\right)$ of the covariance matrix of the observations}
	\KwOut{the generalized estimate $\hat x_\ast$ and the estimated PSF footprint $h$}
	initialize $h$ as a vector of size $\shape a$ with all entries set to $1$\;
	initialize $\hat x_\ast$ using a rough estimate\;
	\Repeat{$m = \left\langle 1, h \right\rangle$}{
		$R\left[k\right] \leftarrow \hat x_\ast^\T M_\ast\, \hat x_\ast$ for all $k$\;
		update $h\!\left[k\right] \gets 0$ where $R\left[k\right]$ is largest\;
		refine the estimate $\hat x_\ast$ using the updated footprint $h$\;
	}
\end{algorithm}

\subsection{Implementation}
\label{chap:method:impl}

The first step to the computation of the estimate $\hat x_\ast$ is the identification of the signal subspace, i.e.\ $\range U$, as described in \autoref{chap:signal_subspace:noisy}. Instead of computing the EVD of the $\shape y \times \shape y$ empirical covariance matrix $\frac{1}{n} YY^\T = U \Lambda U^\T$, that is too large to be kept in memory for high-resolution images, we rely on the truncated \emph{singular value decomposition} (SVD, see, e.g., \cite{murphy2012}) of $Y = V S W^\T$ with $S = \Diag\left(s_1, \dots, s_{\shape a}\right)$ and $s_i \geq s_{i+1}$. We only compute the $\shape a$ largest singular values $s_i$. Since \inline{$\frac{1}{n} YY^\T = \frac{1}{n} V S W^\T W S V^\T = \frac{1}{n} V S^2 V^\T$}, the SVD of $Y$ recovers the eigenvectors $U = V$ of the covariance matrix with corresponding eigenvalues $\lambda_i = s_i^2 / n$.

The SVD implementations which we have considered are shown in \autoref{tab:svd_comparison}. The ``\_getsdd'' routine from LAPACK produces accurate results, but demands that the entire matrix $Y$ is loaded into memory. As the matrix $Y$ becomes too large, we must access it in portions from a slower storage. \citet{halko2011} proposed two SVD implementations with a memory complexity of $\mathcal O\left(\shape y \cdot \left(\shape a + \kappa\right)\right)$, where a greater $\kappa$ increases noise robustness:
\begin{description}
	\item[Single-pass SVD:] The single-pass SVD processes the columns of the matrix $Y$ one-by-one in a streaming fashion. On the downside, the implementation has proven to become inaccurate in the presence of noise.
	\item[Randomized SVD:] The randomized SVD is not designed for off-memory data specifically. Nevertheless, it can be easily adapted for this use-case, since it accesses the matrix $Y$ solely within dot products. With a block-wise dot product implementation, this implementation outperforms the other two in terms of speed at lower resolutions, if the blocks are chosen at least the size of $\shape y$ elements.
\end{description}
To determine the best suited implementation, we have performed a quantitative comparison of the computation time required by the different implementations. The results are shown in \autoref{tab:svd_bechmarks}. %Our experimental Python toolbox offers the choice between the three SVD routines for in-memory data, but only the single-pass SVD for off-memory data.

\begin{table}
	\caption{Qualitative comparison of SVD implementations, based on their memory complexity and noise robustness.}
	\centering
	\begin{tabular}{lll}
		\toprule
		Data size & $\sigma^2/n$ ratio & Suited SVD routine \\
		\midrule\midrule
		
		Small & Any & LAPACK's ``\_gesdd'' \\
		\midrule

		Any & Small & Single-pass SVD \\
		\midrule

		Moderate/high & Moderate/high & Randomized SVD \\
		\bottomrule
	\end{tabular}
	\label{tab:svd_comparison}
\end{table}

\begin{table*}
	\caption{Quantitative comparison of the computation time different SVD implementations on regular consumer hardware. LAPACK was not applicable to very large image data (``---''). The results are reported in seconds and the best results are highlighted.}
	\centering
    \resizebox{\linewidth}{!}{%
	\begin{tabular}{lcccccccc}
		\toprule
		& \multicolumn{4}{c}{$n = 500$} & \multicolumn{4}{c}{$n = 4000$} \\
		\cmidrule(r){2-5}
		\cmidrule(r){6-9}
		& $\shape y = 10^2$ & $\shape y = 10^3$ & $\shape y = 10^4$ & $\shape y = 10^5$
		& $\shape y = 10^2$ & $\shape y = 10^3$ & $\shape y = 10^4$ & $\shape y = 10^5$ \\
		\midrule\midrule
		
		Randomized SVD: & \textbf{$0.011$}
		                          & \textbf{$0.083$}
		                                    & $1.007$ &  $23.523$
		                & \textbf{$0.076$}
		                          & \textbf{$0.577$}
		                                    & $6.977$ & $182.782$ \\
		\midrule
		
		Single-pass SVD: & $0.033$ & $0.112$ & \textbf{$1.000$}
		                                               & \textbf{$11.481$}
		                 & $0.212$ & $0.689$ & \textbf{$4.809$}
		                                               & \textbf{$63.131$} \\
		\midrule
		
		LAPACK: & $0.016$ & $1.858$  & $14.314$ & ---
		        & $0.117$ & $23.534$ & ---      & --- \\
		\bottomrule
	\end{tabular}}
	\label{tab:svd_bechmarks}
\end{table*}

For efficient implementation of the inflating method described in \autoref{chap:method:inflating}, computation of the potentially huge $\shape y \times n \shape d$ matrix $\left(\cnvmat D_1 Y, \dots, \cnvmat D_{\shape d} Y\right)$ should be avoided. The SVD of $Y = USW^\T$ shows that
\begin{gather}
	\range \cnvmat D_t Y = \range \cnvmat D_t US
	\text,
	\label{eq:inflating_impl_eq}
\end{gather}
since $W^\T$ has full column rank. Thus, inflating can be performed efficiently by computing the smaller $\shape y \times m_0 \shape d$ matrix $\left[\cnvmat D_1 US\, \dots\, \cnvmat D_{\shape d} US\right]$, which comes at the cost of an additional SVD. Throughout the results we present in \autoref{chap:results}, we used the LAPACK implementation for the second SVD.

\subsubsection{Rayleigh quotient iterations}
\label{chap:method:impl:rayleigh}

\autoref{alg:estimating_psf_masks} requires the computation of the estimate $\hat x_\ast$ for initialization. Given a roughly known eigenvalue $\mu_0$ of the matrix $M_\ast$, the corresponding eigenvector $\hat x_\ast$ can be computed using \emph{Rayleigh quotient iterations},
\begin{gather}
	\left(M_\ast - \mu_k I\right) \hat x_{\ast, k+1}
	=
	\hat x_{\ast, k},
	\label{eq:rayleigh_quotient_itr:linear_system}
\end{gather}
where $\mu_k = \hat x_{\ast, k}^\T\, M_\ast\, \hat x_{\ast, k}$ for $k \geq 1$. In each iteration, the linear system in Eq.~\eqref{eq:rayleigh_quotient_itr:linear_system} is solved for $\hat x_{\ast, k+1}$ and then normalized. The convergence rate of the iterations is cubic (e.g., \cite{golub:1996}). The choice of $\hat x_{\ast, 0}$ is random.

Since the eigenvector $\hat x_\ast$ corresponds to the eigenvalue of $M_\ast$ which is closest to $0$, choosing $\mu_0 = 0$ is reasonable. We used Newton iterations with Krylov approximation of the inverse Jacobian \citep{kelley1995} for the solution of the linear system.

Algorithm~\ref{alg:rayleigh_quotient_itr} summarizes the approximation of $\hat x_\ast$ to a given precision, which is estimated based upon the convergence of the $\mu_k$ sequence. It is convenient to set the parameter $\alpha$ in Eq.~\eqref{eq:matrix_of_interest_generalized} to $\alpha = 1 / \shape a$, so $\mu_k$ becomes independent of $\shape a$. The bottleneck of the algorithm is the solution of the linear system in Eq.~\eqref{eq:rayleigh_quotient_itr:linear_system}. However, due to the cubic convergence rate of the algorithm, high precisions are reached after only few iterations.

\begin{algorithm}
 	\caption{Computation of the generalized estimate $\hat x_\ast$ using Rayleigh quotient iterations.}
	\label{alg:rayleigh_quotient_itr}
 	
	\SetAlgoLined
	\KwIn{the matrix $M_\ast$ and the required precision $1 / \mu_\Delta$}
	\KwOut{the estimate $\hat x_\ast$}
	initialize $\mu' \leftarrow 0$ and $\hat x_\ast$ randomly\;
	\Repeat{$\left|\mu' - \mu\right| < \mu_\Delta$}{
		$\mu \leftarrow \mu'$\;
		$\hat x_\ast \leftarrow \text{solve } \left(M_\ast - \mu I\right) \hat x'_\ast = \hat x_\ast \text{ for } \hat x'_\ast$\;
		$\hat x_\ast \leftarrow \hat x_\ast / \hat x_\ast^\T \hat x_\ast$\;
		$\mu' \leftarrow \hat x_\ast^\T M_\ast\, \hat x_\ast$\;
	}
\end{algorithm}

We have used used $\mu_\Delta = 10^{-3}$ for \autoref{alg:rayleigh_quotient_itr} in all our experiments.

\subsubsection{Estimate refinements}
\label{chap:method:impl:poweritr}

Recall that besides of computing the estimate $\hat x_\ast$ for initialization, \autoref{alg:estimating_psf_masks} also relies on incremental updating of the estimate. We found that rather rough refinements are sufficient, which can be performed faster than using a single iteration of \autoref{alg:rayleigh_quotient_itr}, as described below.

In \autoref{chap:method:noise_free} we argued that the matrix $M$ is symmetric and positive semidefinite, and so are the matrices $Q\!\left(h\right)$ and $M_\ast$ in Eq.~\eqref{eq:matrix_of_interest_generalized}. Let $\mu_\text{up}$ be an upper bound of the eigenvalues of the matrix $M_\ast$ and let $\mu_\text{min}$ be the smallest eigenvalue. We define the spectrum-shifted matrix $Z = \mu_\text{up} I - M_\ast$ and observe that $Z \hat x_\ast = \mu_\text{up} \hat x_\ast - \mu_\text{min} \hat x_\ast$, so $\hat x_\ast$ is an eigenvector of matrix $Z$ with eigenvalue $\mu_\text{up} - \mu_\text{min}$, which also is the largest eigenvalue of the matrix $Z$.

For $k\to\infty$, the \emph{power iterations} $\hat x_{\ast,k+1} = Z \hat x_{\ast,k}$ recover the eigenvector $\hat x_{\ast}$ of the matrix $Z$ corresponding to the largest eigenvalue of $Z$ (see, e.g., \cite{golub:1996}). For the upper bound $\mu_\text{up}$ of the eigenvalues of the matrix $M_\ast$ we used $\mu_\text{up} = 1 + \alpha \shape a$, which is legitimate due to the following two reasons:
\begin{enumerate}
	\item The largest eigenvalue of matrix $M_k$ in Eq.~\eqref{eq:matrix_of_interest_generalized} is $1$, the matrix since $\cnvmat B_k$ contains at most a single $1$ in each of column, and $I - UU^\T$ is a projection matrix.
	\item The matrix $Q\left(h\right)$ is a binary diagonal matrix, with eigenvalues $0$ and $1$.
\end{enumerate}
\autoref{alg:power_itr} summarizes the resulting procedure for the refinement of the estimate $\hat x$.

\begin{algorithm}
 	\caption{Refinement of the generalized estimate $\hat x_\ast$ using power iterations.}
	\label{alg:power_itr}
 	
	\SetAlgoLined
	\KwIn{$\shape a$, the matrix $M_\ast$, an initial estimate $\hat x_\ast$, and the required precision $1 / \mu_\Delta$}
	\KwOut{the refined estimate $\hat x_\ast$}
	define $Z \leftarrow \left(1 + \alpha \shape a\right) I - M_\ast$\;
	initialize $\mu \leftarrow \hat x_\ast^\T Z\, \hat x_\ast$ and $w \leftarrow \hat x$\;
	\Repeat{$\left|\mu' - \mu\right| < \mu_\Delta$}{
		$\mu'         \leftarrow \mu$\;
		$\hat x_\ast  \leftarrow w / w^\T w$\;
		$w            \leftarrow Z \hat x_\ast$\;
		$\mu          \leftarrow x_\ast^\T w$\;
	}
\end{algorithm}

We have used used $\mu_\Delta = 10^{-4}$ for \autoref{alg:power_itr} in all our experiments.

\subsubsection{Further optimizations}
\label{chap:method:impl:optimizations}

Instead of using \autoref{alg:estimating_psf_masks} to jointly compute the generalized estimate $\hat x_\ast$ and the PSF footprint, a two-step scheme is more efficient. In the first step, the algorithm is used for only a small image section of the observations to compute the PSF footprint. In our experiments, using image sections of width and height $3$ to $8$ times larger than $\shape a$ offered a good trade-off between reliability and speed. In the second step, $\hat x_\ast$ is computed directly using \autoref{alg:rayleigh_quotient_itr} and the already determined PSF footprint.

\autoref{alg:estimating_psf_masks}, \ref{alg:rayleigh_quotient_itr}, and \ref{alg:power_itr} require the $\shape x \times \shape x$ matrix $M_\ast$ as input. However, since this matrix only appears within dot products, there is no need to compute its explicit representation. We rewrite the linear system $\left(M_\ast - \mu I\right) \hat x_\ast' = \hat x_\ast$ in Algorithm~\ref{alg:rayleigh_quotient_itr} as $M_\ast \hat x_\ast' - \mu \hat x_\ast' - \hat x_\ast = 0$ for this purpose. Therefore, the memory complexity of the three algorithms is not larger than $\mathcal O\left(\shape x + m \shape y\right)$.

The computation of dot products with the matrix $M_\ast$ is a very frequent operation, which is worth to be implemented for maximum efficiency. To this end, we implemented the matrix-vector-product $\cnvmat B_k x$ in Eq.~\eqref{eq:matrix_of_interest_generalized} as image cropping, and $\cnvmat B_k^\T y$ as image padding operations. Depending on the implementation of the linear algebra, the batched computation of
\begin{gather}
	UU^\T \left(\dots, \cnvmat B_k x, \dots\right)
\end{gather}
may be faster than the sequential
\begin{gather}
	\dots, UU^\T \cnvmat B_k x, \dots
	\text,
\end{gather}
as it allows to exploit memory localities. In our Python-based implementation, this optimization accelerated the dot product computations by up to factor $7$. Furthermore, it is appropriate to evaluate the batches in parallel.

For post-processing, we scaled $\hat x_\ast$ by a factor determined as the mean pixel value of the observations.

\subsection{Experimental results}
\label{chap:results}

In this section, we describe the application of our eigenvector-based method to synthetically generated image sequences. The pixel values were of all images were restricted to the interval $\left[0, 255\right]$. All experiments were performed using dated consumer hardware, comprising only 4 GiB RAM and an Intel Core i5-3320M CPU. In addition, we have also included a quantitative comparison of our method to \cite{harmeling2009}.

For quantitative evaluation of the results, we use the norm-invariant root mean square (RMS) error,
\begin{gather}
	\frac{1}{\sqrt{\shape x}} \cdot \left\|x_\text{true} - \frac{\left\|x_\text{true}\right\|}{\left\|x\right\|} \cdot x \right\|_\mathsf{Fro},
	\label{eq:ni_rms}
\end{gather}
which is invariant to the norm of the result image $\hat x_\ast$. We ignore those pixels of $\hat x_\ast$ and $x_\text{true}$, which were not observed due to the generated PSFs (cf.\ \autoref{chap:when_inflating_fails}).

\subsubsection{Moderate noise levels}
\label{chap:results:n_hi}

\begin{figure}
	\centering
	\subfloat[PSF footprint]{
        \begin{minipage}[t]{3cm}
            \includegraphics[width=\linewidth]{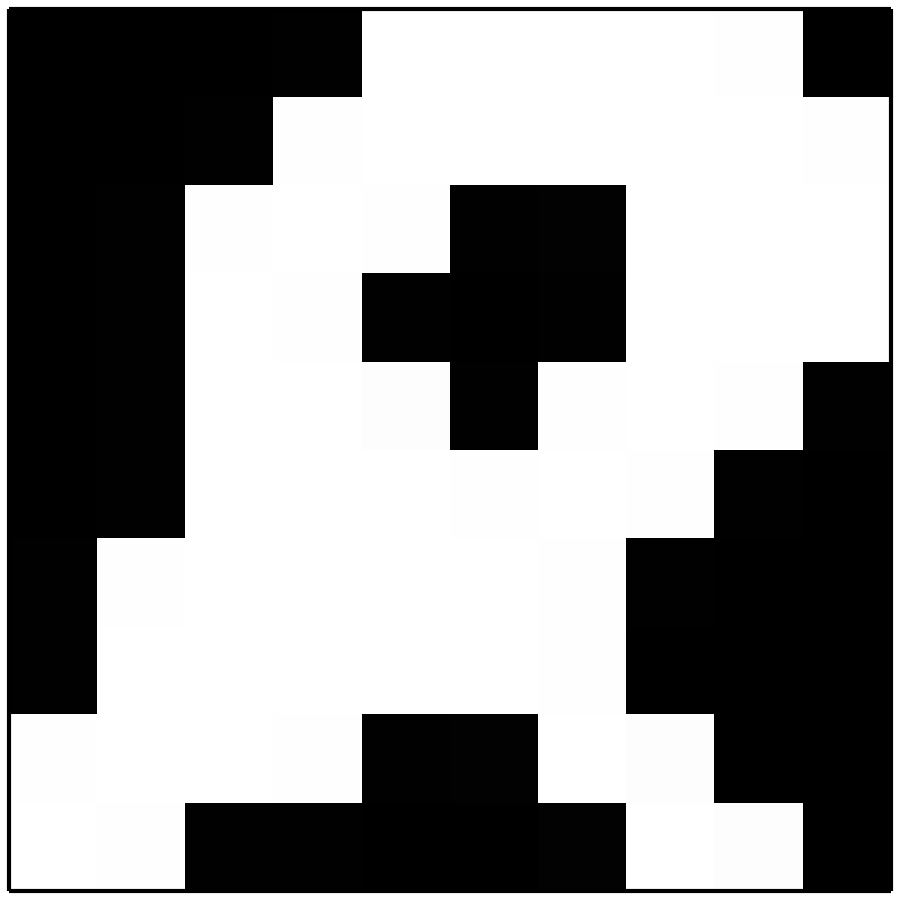}
        \end{minipage}
    	\label{fig:results:psf_mask}}
	\subfloat[$\sigma^2 = 10^{-1}$]{
        \begin{minipage}[t]{3cm}
    		\includegraphics[width=\linewidth]{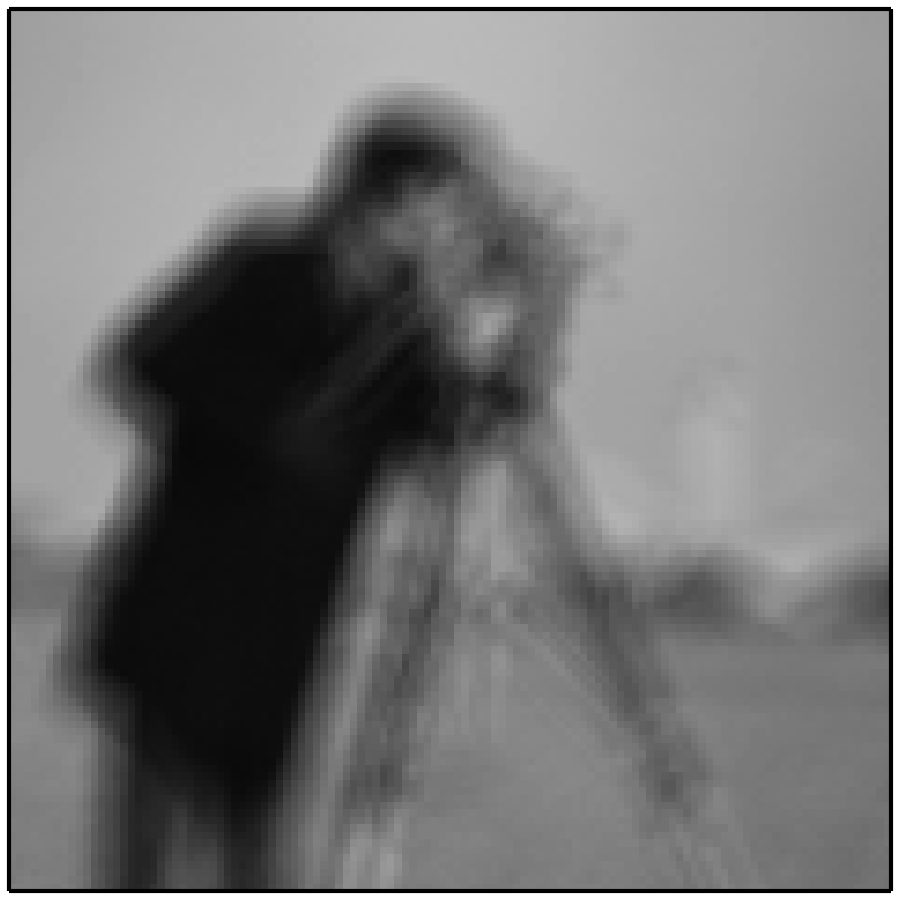}\\
            \includegraphics[width=\linewidth]{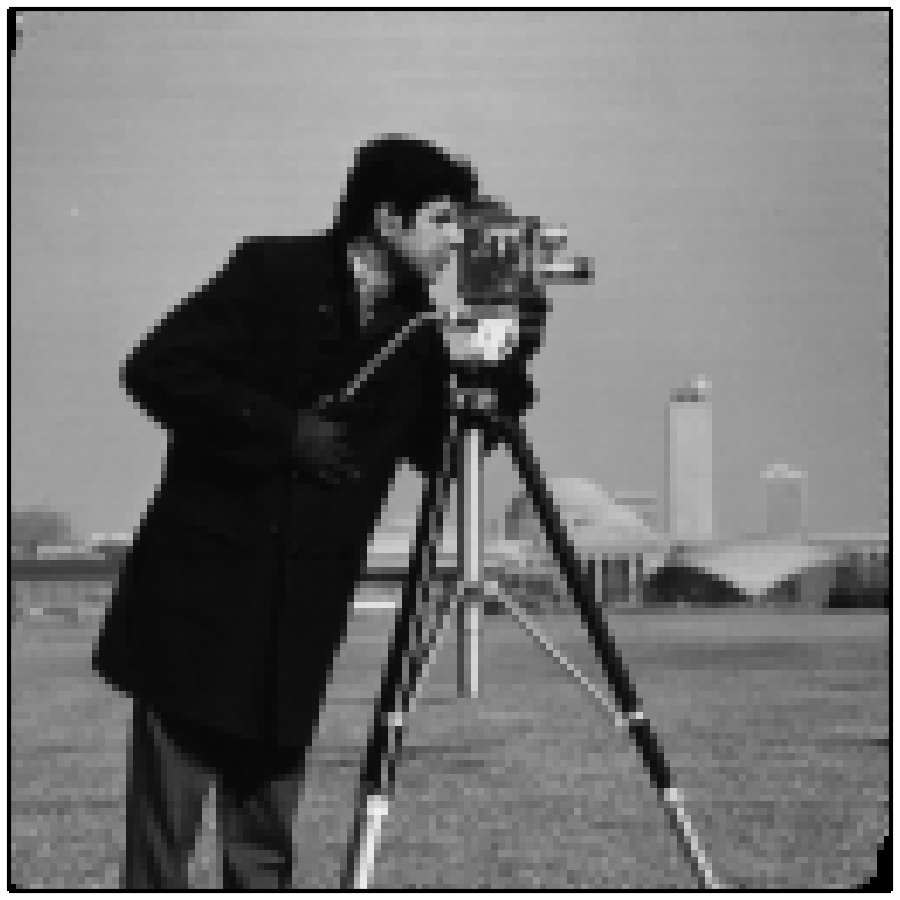}
        \end{minipage}
    	\label{fig:results:hi_n:y1:lo_noise}}
	\subfloat[$\sigma^2 = 10^{0}$]{
        \begin{minipage}[t]{3cm}
    		\includegraphics[width=\linewidth]{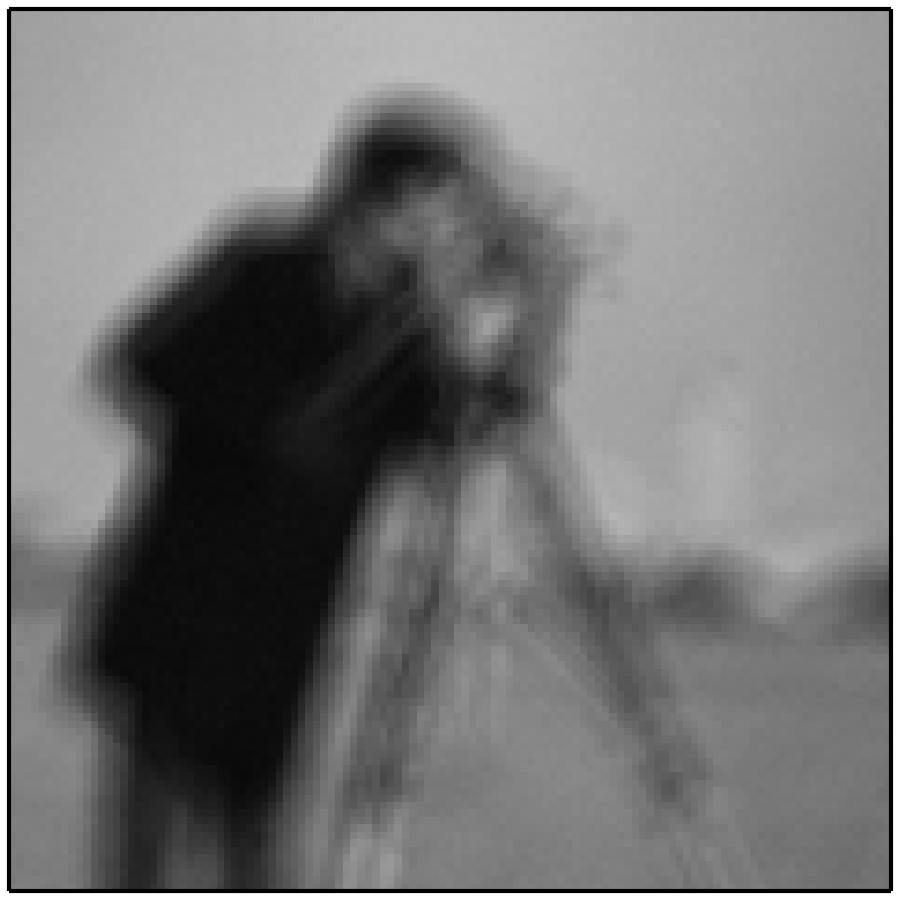}\\
            \includegraphics[width=\linewidth]{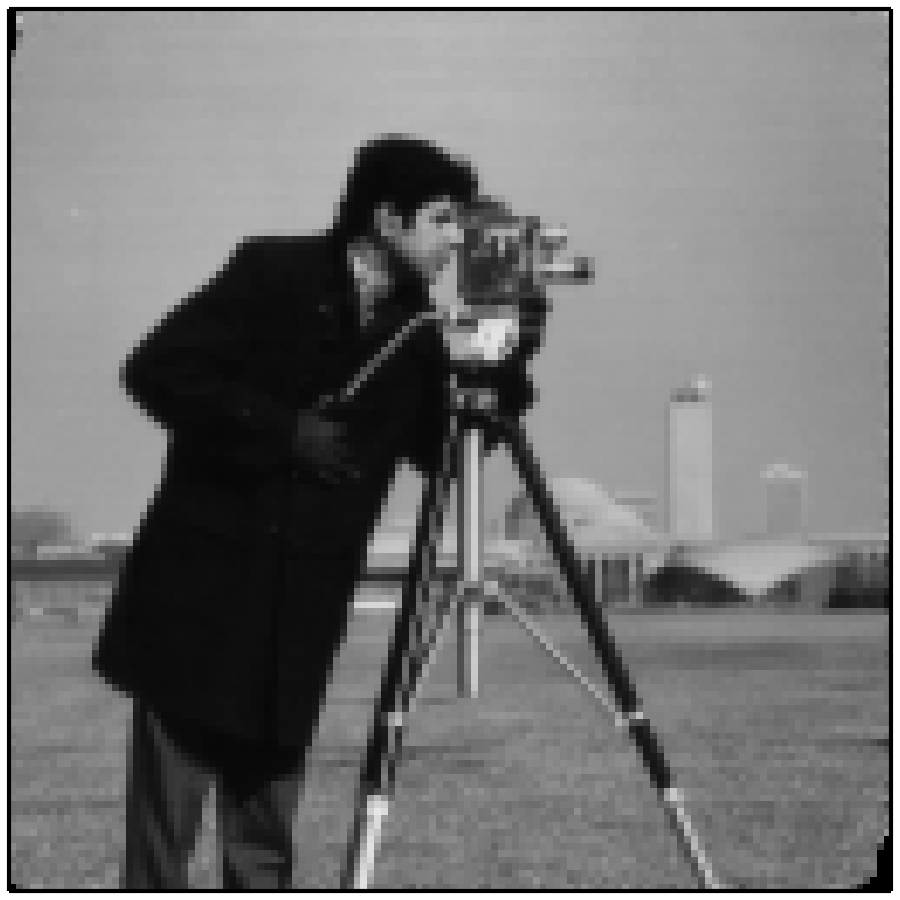}
        \end{minipage}
    	\label{fig:results:hi_n:y1:mi_noise}}
	\subfloat[$\sigma^2 = 10^{1}$]{
        \begin{minipage}[t]{3cm}
    		\includegraphics[width=\linewidth]{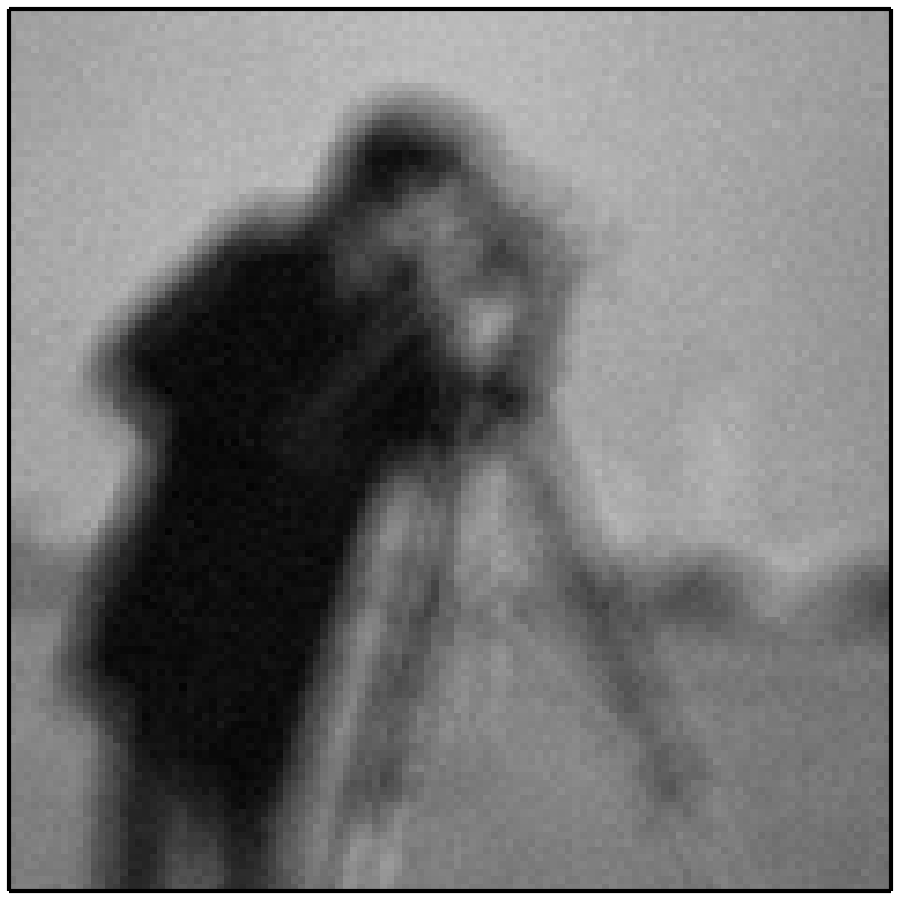}\\
            \includegraphics[width=\linewidth]{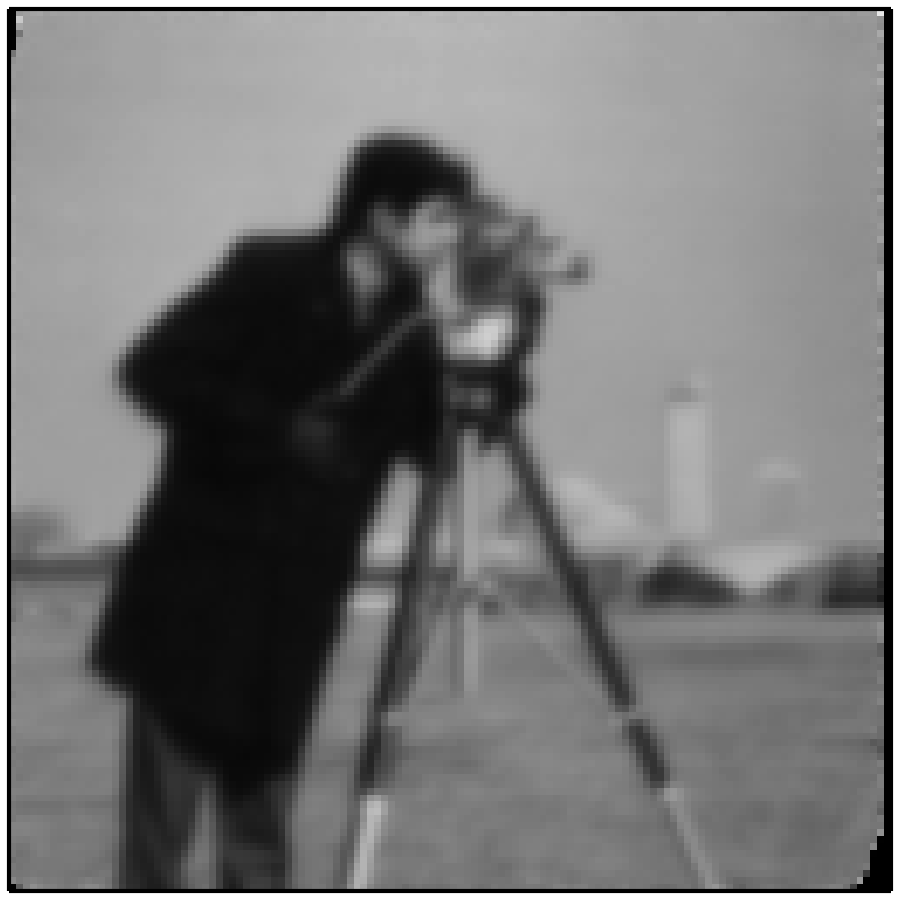}
        \end{minipage}
    	\label{fig:results:hi_n:y1:hi_noise}}
	\caption{(a) $10 \times 10$ pixels PSF footprint, where black pixels correspond to 0 and white pixels correspond to 1, and (first row) synthetically generated example observations, generated using the PSF footprint at different noise levels (b)--(d). (second row) The corresponding results obtained using the proposed eigenvector-based MFBD method.}
	\label{fig:results:moderate_noise}
\end{figure}

In a first experiment, we have used image sequences comprising $n = 1000$ images, each sequence generated using an individual noise level $\sigma^2 \in \left\{0.1, 1, 10\right\}$. The images were generated according to Eq.~\eqref{eq:model} using the ground truth image from \autoref{fig:intro:cameraman} of size $\shape x = 128 \times 128$ pixels and random PSFs of size $\shape{a,0} = 10 \times 10$ pixels, while only varying the $57$ pixels of the PSF footprint shown in \autoref{fig:results:psf_mask}. Example images from the three sequences are shown in \autoref{fig:results:hi_n:y1:lo_noise}--\ref{fig:results:hi_n:y1:hi_noise}.

The first step of our method is the recovery of the matrix $U$ which represents the signal subspace. For the lower noise levels $\sigma^2 = 0.1$ and $\sigma^2 = 1$, we have used the randomized SVD \cite{halko2011} to compute the EVD of the covariance matrix $\frac{1}{n} YY^\T$. In this experiment, we know that the dimension of the PSF subspace is equal to the number of the 1-entries in the PSF footprint, thus we skip inflating and proceed to the estimation of the PSF footprint directly. Using centered image sections of the size $\shape y' = 75 \times 75$ pixels, we exactly recovered the PSF footprint as described in \autoref{chap:method:impl:optimizations}.

The case correpsonding to the increased noise level $\sigma^2 = 10$ is more challenging. Since the obtained matrix representation of the signal subspace of the signal subspace is less accurate, the estimation of the PSF footprint yields inaccurate results. Fortunately, inflating the observations turns out being helpful. The reason is that, in contrast to the estimation of the PSF footprint, inflating reduces the gap between the signal subspace dimension $m$ and $\shape a$ in a non-heuristic manner. We then estimated the PSF footprint from centered image sections of the size $\shape y' = 38 \times 38$ pixels of the inflated observations.

The final step of our method concerns the computation of the estimate $\hat x_\ast$ using the priorly estimated PSF footprint. The results are shown in the bottom row of \autoref{fig:results:moderate_noise}. For $\sigma^2 = 0.1$, the undeteriorated image is recovered almost perfectly up to the unobserved pixels in the upper left und lower right corners of the image (RMS value of 2.63). The result is somewhat blurrier for $\sigma^2 = 1$ (RMS value of 5.05). The result obtained for $\sigma^2 = 10$ is even blurrier (RMS value of 9.97 and 17.69 without inflating). This could be improved by taking more observations into account. In all three cases, the overall runtime of our method was about 1 minute without inflating, and increased to about 3 minutes using the inflated observations.

\subsubsection{High noise levels}
\label{chap:results:comparison}

\begin{figure}
	\centering
    \subfloat[]{\includegraphics[height=40mm]{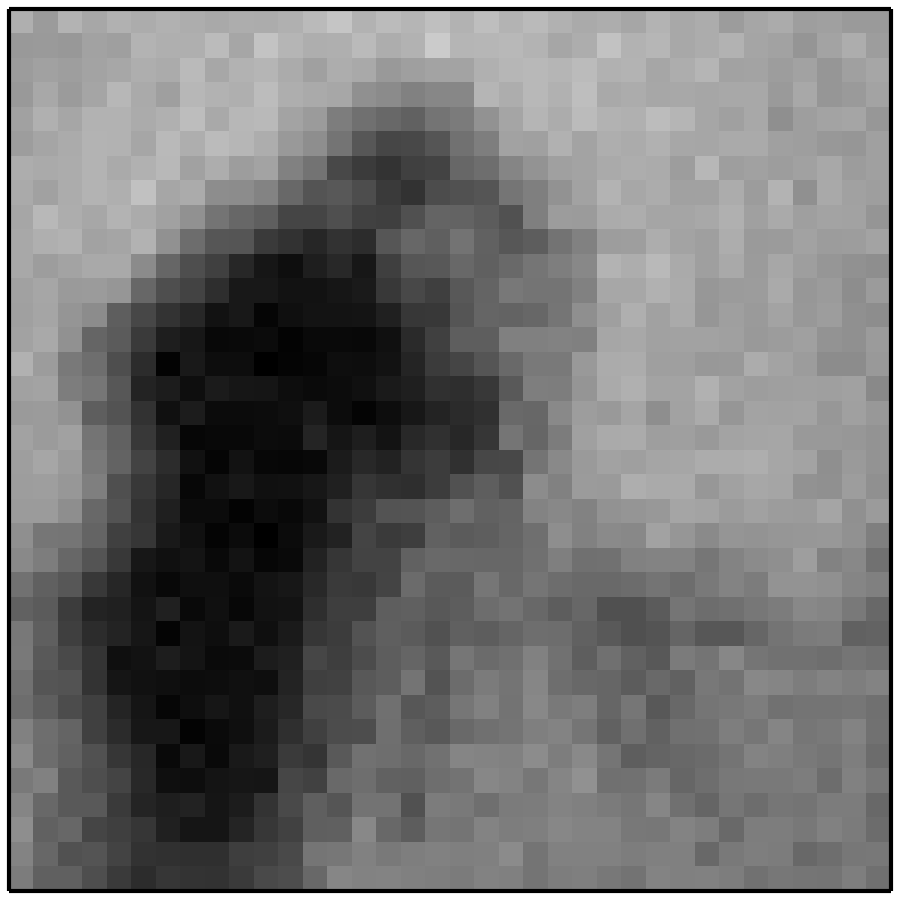}\label{fig:results:high_noise:y1}}
    \quad
	\subfloat[]{\includegraphics[height=40mm]{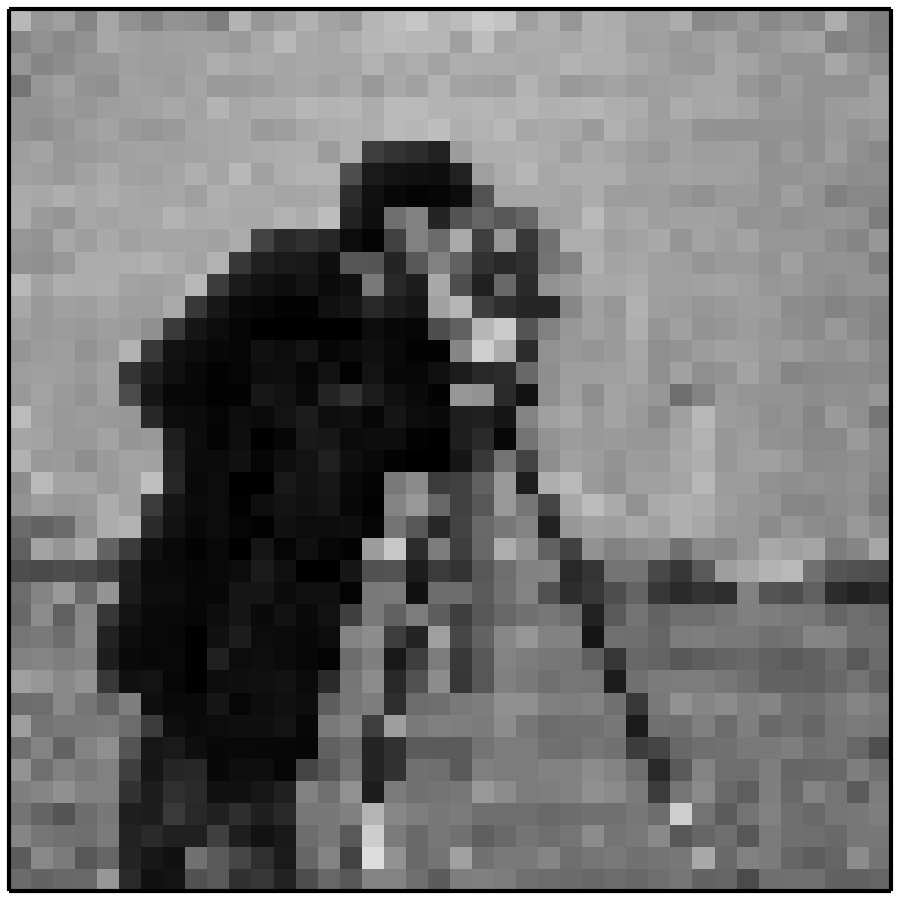}\label{fig:results:high_noise:obd}} % After $9.3$ seconds
    \quad
    \subfloat[]{\includegraphics[height=40mm]{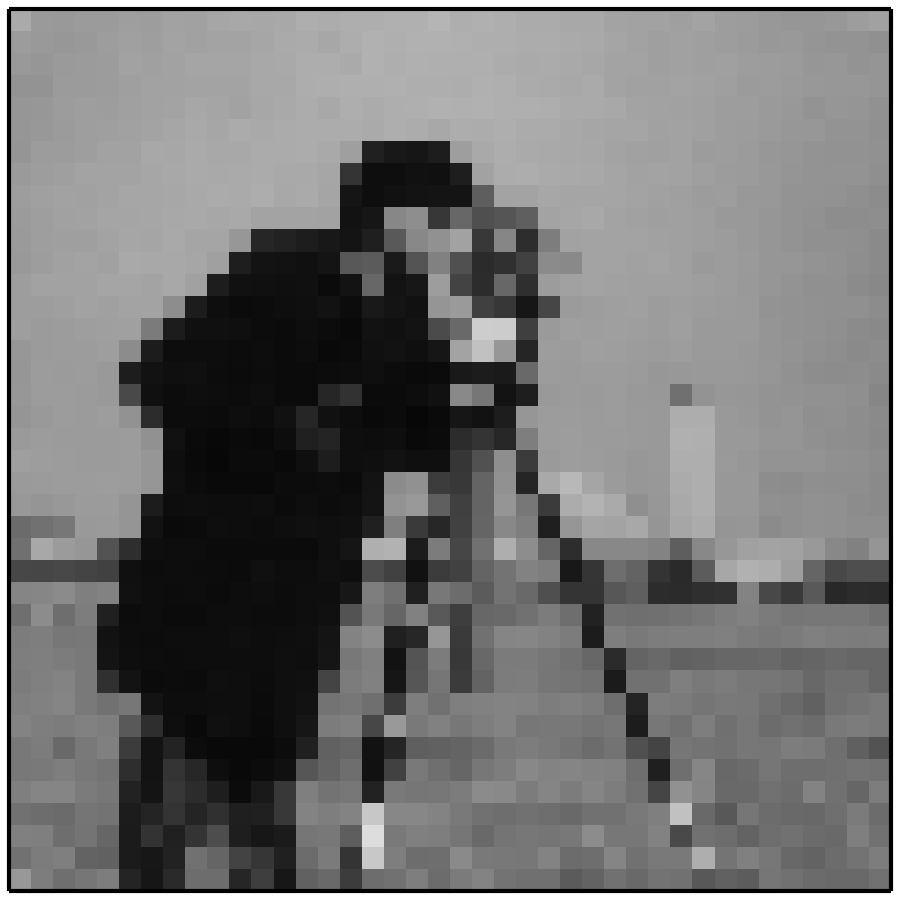}\label{fig:results:high_noise:ours}}
	\caption{(a)~Example observation generated using the high noise level $\sigma^2 = 50$, (b)~result obtained using \cite{harmeling2009}, and (c)~result obtained using the proposed eigenvector-based MFBD method.}
	\label{fig:results:high_noise}
\end{figure}

Our method specifically addresses the case of very high noise levels and large numbers of observations $n$. Thus, in a second experiment, we have used a larger image sequence comprising $n = 5000$ images, which was generated using the noise level $\sigma^2 = 50$. We also reduced the ground truth size to $\shape x = 40 \times 40$ pixels and used PSFs of the smaller size $\shape a = 5 \times 5$ pixels. \autoref{fig:results:high_noise:y1} shows an example image from the sequence.

In \autoref{chap:related_work}, we described that other methods like \cite{harikumar1999} and \cite{sroubek2012} are intractable for this case due to the large size of the involved matrices. For example, the latter demands computation of a matrix of about 24 GiB using single-precision floating point numbers. For this reason, we compare our method against the online method \cite{harmeling2009}. However, this method is originally based on circular convolution. We found that, if implemented using valid convolution instead, the online method fails to converge if $m < \shape a$. We thus used $m = \shape a$ to generate the image sequence for this experiment. The result obtained using the online method yields an RMS value of 7.58. The method took 245 seconds to process the whole image sequence.

For our method, we again used the randomized SVD \cite{halko2011} for the recovery of the signal subspace, which terminated after $8.7$ seconds. Due to $m = \shape a$, neither inflating needs to be performed, nor do we need to estimate the PSF footprint. The subsequent computation of the estimate $\hat x$ using \autoref{alg:rayleigh_quotient_itr} took $0.5$ seconds. Both steps took only $9.3$ seconds in total and the obtained result yields an RMS value of 5.67. For comparison, we also computed the result obtained using the online method after $9.3$ seconds, which corresponds to an RMS value of 12.49. The results are shown in \autoref{fig:results:high_noise:obd}--\ref{fig:results:high_noise:ours}. It can be seen that the result obtained using our method is minorly sharper than the result obtained using \cite{harmeling2009} and far less noisy.

Overall, our method yields a significantly improved result using the same computation time (RMS value of 12.49 compared to 5.67), and also an improved result if the online method is given more computation time (RMS value of 7.58 compared to 5.67).

\section{Conclusions and future work}
\label{chap:discussion}

We have presented two methods for multi-frame blind deconvolution method, which recover an undeteriorated image from a sequence of its blurry and noisy observations. This is accomplished by exploiting the signal subspace, which is encoded in the empirical covariance matrix of the observations. The first presented method is based on likelihood maximization and requires careful initialization to cope with the non-convexity of the loss function. The second presented method circumvents this requirement by exploiting that, under two specific conditions, the same solution also emerges as an eigenvector of a specifically constructed matrix. The matrix is fully determined solely by the observations, so the filters corresponding to the observations do not need to be estimated, so alternating optimization schemes are not required. We have applied the eigenvector method to synthetically generated image sequences and performed a quantitative comparison with a previous method, obtaining strongly improved results.

The first condition demands that the number of observations is sufficiently large, so that the signal subspace of the noisy observations approximates the signal subspace of the unobservable, noise-free observations. The second condition demands that the dimension of the signal subspace is sufficiently large. To cope with this, we have described a pre-processing step which \emph{inflates} the signal subspace by artificially generating additional observations. In addition, we have proposed an extension of the eigenvector method which copes with insufficient dimensions of the signal subspace by estimating a \emph{footprint} of the unknown filters using an alternating optimization scheme.

Application of the proposed methods to a large variety of image data will be subject of future work. This will particularly comprise high-resolution and real-world images, as well as a more comprehensive evaluation, including more previous methods for comparison. Stable implementations of the proposed methods should automatically choose the best-suited method for computation of the SVD. Interesting open research questions were also pointed out in \autoref{chap:method:noisy_ideal}.

%Bibliography
%\bibliographystyle{unsrt}
%\bibliographystyle{plainnat}
\bibliographystyle{unsrtnat}
\bibliography{references}

\end{document}